\documentclass[10pt,twocolumn]{article}
\usepackage[letterpaper,margin=.75in,columnsep=.25in]{geometry}
\usepackage[T1]{fontenc}
\usepackage{lmodern,amsmath,amssymb,booktabs,graphicx,microtype}
\usepackage[hidelinks]{hyperref}

\newcommand{\KL}{\operatorname{KL}}
\newcommand{\clip}{\operatorname{clip}}
\title{What Transfers from Training-Only Priors?\\Separating Knowledge, Ordinal Targets, and Optimization}
\author{Qiye Lu \quad Jiang Ji \quad Liang Zhang\\[.3em]\small Ant International}
\date{23 October 2026}
\begin{document}
\maketitle
\begin{abstract}
Training-only teacher access lets a compact student deploy with local
features and exported weights. This established distillation interface does
not identify what transfers or when it helps. We separate label, query,
capacity, and source-access budgets on three public credit archives and a
previously unexamined bankruptcy archive. Shared tree parameters, shuffles,
and optimizer interventions expose limitations of historical Ranking Prior
Alignment; no tested single repair consistently rescues it. Heldout source
queries and matched warmup and mappings retain a rank-listwise advantage
over mapped staged KD in 24 credit configuration means and all six
teacher--capacity means of a frozen new-archive test, with explicit query
failures and stronger source-weight delivery under a different interface.
A pretrained TabICLv2 follow-up at 100 labels initially improves matched
supervision in only one of eight means; six of nine source-predicted policy
gains reverse on target. A registered seen-archive intervention adds 500
or 2,000 unlabeled queries: all eight matched AUC means become positive
under selected and fixed alpha. However, outer effects are mixed and no
conditional paired-draw interval for these supervised contrasts lies wholly
above zero. Increasing teacher context at 500 student labels does not beat
matched supervision in any of eight ordinary-context means, despite stronger
direct teacher performance. Rank KD retains substantial probability errors.
Independent reloads verify zero guarded external inference accesses. Results
support query/context boundary diagnostics, without universal superiority,
population significance, a prospective transfer rule, or an LLM-specific
advantage.
\end{abstract}

\paragraph{Scope of this revision.}
This revision extends the \href{https://arxiv.org/abs/2610.11146v2}{earlier
ranking-prior study} with matched distillation and optimizer controls,
heldout training queries, a new-archive test, and pretrained-teacher
query/context interventions. It clarifies the implemented RPA surrogate
and qualifies earlier interpretations concerning granularity and
sample--capacity rules. Historical configurations remain bounded findings;
the added controls do not establish general RPA superiority or a validated
prospective deployment rule.

\section{Introduction}
Credit-scoring systems may have mature knowledge about past customers but
few labels for a new population. Calling an external predictor in production
can conflict with latency requirements. Our interface therefore restricts
teacher access to training: the deployed predictor uses its own weights and
available features, with zero external calls. This interface is shared by
knowledge distillation~\cite{hinton2015}; we do not claim it as a new paradigm.
The target-label budget $N$, training-query budget $N+M$, and student
resource budget are distinct constraints. Restricting inference calls does
not require restricting training queries to labeled records. Weight delivery
provides another route when the source provider can export a small model.

The research problem is to determine what an apparent improvement means.
A method may use useful teacher information yet lose to a simpler method.
It may also benefit from different gradients, curvature, or initialization
without transferring the teacher's record-level correspondence. Empirical
ranks offer an inexpensive alternative to raw probability targets, but
invariance to monotone transformations does not itself establish superiority.

We revisit Ranking Prior Alignment, a cross-record listwise update related
to learning-to-rank~\cite{cao2007}. We ask three questions: does teacher-score
correspondence improve the student; how does the representation compare with
pointwise distillation; and which observed sample-size trends arise from the
optimizer interface? Attention-based MIL~\cite{ilse2018} provides a second
mechanism question: whether a risk score is an appropriate relevance target.

Our current contributions are a public protocol with explicit information
budgets, matched distillation and optimizer controls, and a diagnosis of
transfer boundaries. An algorithm advantage must survive these controls.
We additionally test multiple teacher families, stronger target learners,
and a separate source-weight delivery interface. Heldout-query experiments
separate adding unlabeled queries from source alpha reselection, match
the pointwise and listwise warmup and teacher representation, and test
probability offsets without spending additional target labels. The
pretrained-teacher follow-up then intervenes on query coverage and teacher
context while explicitly accounting for the student-label budget.

\begin{figure*}[!t]
\centering
\includegraphics[width=\textwidth]{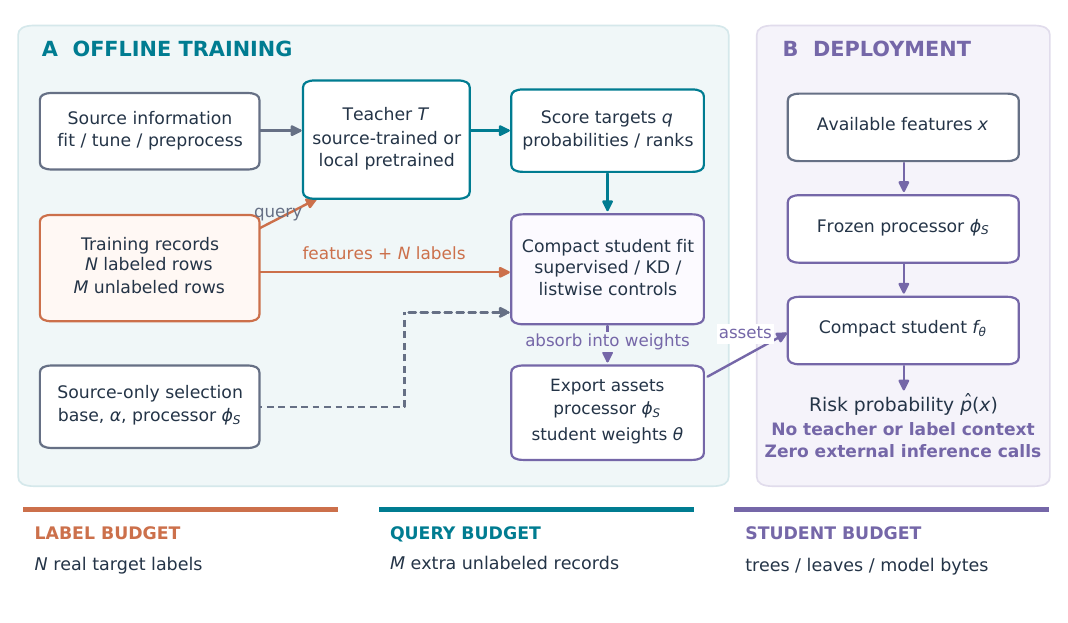}
\caption{The established training-only teacher interface studied here.
Offline access supplies scores and source-derived choices; student fitting
exports only a frozen source processor and compact weights. The production
path takes available features and makes no teacher or external inference
call. Source fitting and development labels are additional information
beyond the target-label budget $N$. The $M$ additional queried records do
not imply $M$ teacher row predictions when several teachers score each row.
In the pretrained experiments, OOF contexts exclude each scored row's label.
The diagram is an interface, not a new network architecture or a claim that
distillation always helps. Direct local ICL also avoids external calls but
retains pretrained weights and labeled context at inference.}
\label{fig:training-deployment}
\end{figure*}

Figure~\ref{fig:training-deployment} makes the inference boundary explicit.
The following evidence separates optimizer behavior
(Figure~\ref{fig:rpa-diagnosis}), query and context interventions
(Figure~\ref{fig:query-context-boundaries}), and deployed asset cost from
predictive quality (Figure~\ref{fig:deployment-quality}).

\section{Setting and Competing Objectives}
The source developer uses labeled records $D_S$ to fit teacher $T$, feature
vocabulary, and hyperparameters. The target student receives $N$ labeled
records $D_T^N$ and scores $q_i=T(x_i)$ on those records. Source labels are
not directly passed to the target learner; source-derived choices are still
additional information and are disclosed. At inference, $f_\theta(x)$ uses
neither teacher scores nor teacher-generated features.

\paragraph{What ``internalization'' establishes.}
Operationally, teacher information is internalized when training changes
the exported student parameters and the deployed map is
$\hat p(x)=\sigma(f_\theta(\phi_S(x)))$, where $\phi_S$ is a frozen
source-fitted preprocessing map. This definition concerns dependencies,
not semantic understanding. Export equivalence establishes an independent
inference path; an informative--shuffle comparison tests record
correspondence; an informative--supervised or informative--KD comparison
tests practical value. None of these alone establishes the other two.
Source fitting, development labels, and source-derived choices remain
part of the information cost even when the target uses only $N$ labels.

Providing source model weights is a distinct feasible interface. We
separately examine direct deployment of a source-trained small model and
target adaptation of that model, both without external calls. These
references require source weight delivery, whereas our main score-only
comparison receives scores on training records. A larger source teacher
could also be hosted locally if its weights and resource budget permit;
zero external calls alone does not require distillation.

\paragraph{Distillation.}
For raw KD, a native logistic tree objective receives
$t_i=(1-\alpha)y_i+\alpha q_i$. For rank KD, replace $q_i$ with
$r_i=\operatorname{rank}(q_i)/N$, using average ranks at ties.
All native rounds receive these targets, including target-dependent
intercept initialization. A staged KD arm instead shares five-round
hard-label warmup followed by the remaining alignment rounds. Rank targets are useful
ordinal supervision, not calibrated default probabilities. Brier and logloss
are retained in the result artifacts; no target-test calibration is fitted.

\paragraph{Historical listwise surrogate.}
Let $f_i$ be student margins, $u_i$ their min--max-normalized values, and
$v_i$ similarly normalized teacher scores. Define
$P_i=\operatorname{softmax}(u/\tau)_i$ and
$Q_i=\operatorname{softmax}(v/\tau)_i$.
The implementation passes the following quantities to XGBoost~\cite{chen2016}:
\begin{align}
g_i^{R}&=\gamma\clip\left(\frac{P_i-Q_i}{\tau\max(\Delta_f,10^{-6})},-1,1\right),\\
h_i^{R}&=\gamma\clip\left(\frac{P_i(1-P_i)}{\tau^2\max(\Delta_f,10^{-6})^2},\epsilon,5\right),
\end{align}
where $\Delta_f=\max f-\min f$. These are added to logistic task
gradient and curvature. Min--max extrema are treated as stopped gradients;
the curvature is diagonal and clipped. This is a surrogate update, not the
full derivative and Hessian of live-normalized KL. The short configuration
performs five warmup rounds and a single ten-round alignment continuation
with constant $\gamma$; it does not anneal.

\paragraph{Separate effects.}
For test metric $A$, we distinguish RPA minus label-only, RPA minus its
own shuffled teacher scores, RPA minus matched curvature-only, and RPA
minus rank KD. Shuffle keeps the score multiset and breaks correspondence;
it can itself harm optimization. Therefore a larger teacher--shuffle gap is
not equivalent to larger deployable AUC. Curvature-only sets $g^R=0$
while retaining $h^R$, which does not depend on teacher values at a fixed
student state. Later states can differ across arms.

\paragraph{Size normalization hypothesis.}
Replicating a score distribution $k$ times divides each softmax probability
by $k$. Away from clipping, each ranking gradient shrinks by $1/k$ while
the logistic task gradient is unchanged. Using
$\gamma_N=\gamma_{100}N/100$ compensates this mechanical gradient scale;
with a fixed curvature floor, it simultaneously increases teacher-free
curvature. We therefore cross fixed/normalized weights with
$\epsilon\in\{.01,10^{-12}\}$, each with its own shuffle and curvature
control. This is a diagnostic intervention, not a proposed universally
better optimizer.

\paragraph{Alternatives without live min--max.}
For a fixed-temperature rank-listwise candidate, define
$z_i=4(r_i-.5)$, $Q=\operatorname{softmax}(z)$, and
$P=\operatorname{softmax}(f)$ at temperature one. Its loss is
$L_L=N\KL(Q\|P)$, with gradient $N(P_i-Q_i)$ and diagonal
curvature $NP_i(1-P_i)$. The sample factor matches the task's sum-logloss
scale; no margin min--max or nonnumerical clipping is used. A raw-logit
variant uses $z_i=\operatorname{logit}(q_i)$ with numerical endpoint
protection only. These are established listwise constructions, not a
claim of a new objective family.

A soft pairwise alternative uses teacher targets
$s_{ij}=\sigma(z_i-z_j)$ and student margins $f_i-f_j$:
\[
L_P=\frac{1}{N-1}\sum_{i<j}\operatorname{BCE}
\bigl(s_{ij},\sigma(f_i-f_j)\bigr).
\]
We use its analytic gradient and Hessian diagonal, not its full coupled
Hessian. Both candidates mix $(1-\alpha)L_{\rm task}+\alpha L_{\rm prior}$
after five hard-label warmup rounds. Their derivatives are checked by
finite differences. Pairs do not create additional labels or independent
evaluation units.

\paragraph{Probability identifiability.}
For fixed $Q$, $L_L(f+c\mathbf 1)=L_L(f)$ for any constant $c$.
Consequently, a pure listwise objective cannot identify the overall
probability level of $\sigma(f)$. Its exact Hessian is
$N(\operatorname{diag}(P)-PP^\top)$ and annihilates $\mathbf 1$;
the diagonal approximation generally does not preserve this null space.
With $\alpha=1$, the alignment phase has no hard-label term to anchor
the offset, although its initialization contains label warmup. This is
an objective-level identification fact, not a sufficient causal explanation
of a particular Brier-score change. Rank targets likewise are not
conditional default probabilities. We therefore evaluate discrimination
and probability errors separately; calibration cannot use test labels.

\section{Public Experimental Protocol}
\paragraph{Data and information access.}
LendingClub~\cite{lendingclub} source fitting uses 2012--2014, target
training uses 2015, and evaluation uses 2016 and 2017 separately. Source
teacher fitting consumes 50,000 labels, disjoint source development 4,000,
and three 100-record source tuning draws. The archive retains eventual
Fully Paid/Charged Off loans, so this is retrospective cohort transfer,
not an as-of lending backtest.

Amex~\cite{amex} uses customer-level S/B numeric statement means with
fixed zero filling and a predeclared 25,000-customer public subset. Its
earlier per-outer partitions contain 12,000 source-side customers, 6,000 holdout
customers, and the remainder as a target pool; the latest protocol
additionally reserves an independent source query pool. Source teacher fitting uses
10,000 labels and development 1,700; three 100-record source tuning draws
sample the remaining source pool and may overlap. This is within-archive
simulated scarcity, not cross-institution generalization. The all-allowed
feature arm is the cached S/B vocabulary, not every raw Amex field.

UCI Credit~\cite{uci} uses fixed original numeric fields with ID excluded,
customer-level partitions, and the same source-side budgets as Amex.
Six monthly statements are instances of one customer, not independent
observations or separate temporal cohorts. The official archive, its checksum,
fixed field mapping, and completed runs are recorded in the execution log.

All tasks use the same available vocabulary for the primary teacher and
student. Earlier controls compare all allowed numeric fields with a diagnostic weak
vocabulary selected using source labels only. Whole-archive categorical
vocabularies are excluded from LendingClub. Numeric mappings are fixed;
no transformation is fitted to target-test data. Weak-feature results alone
are insufficient evidence for deployment coverage.

\paragraph{Heldout source-query protocol.}
The latest follow-up freezes a new namespace before training on the same
three previously examined archives. Each outer adds a separate 4,000-record
source unlabeled pool, disjoint from teacher fitting, source development,
and the 300-record source tuning pool. Its labels are not used. All source,
target, and test groups are disjoint within each outer. Amex retains 3,000
target customers and UCI 8,000 after allocating this pool; LC retains its
2015 target cohort. A source-fit-only processor imputes by medians, clips
at .5\%/99.5\% quantiles, and standardizes features. Earlier protocols and
their results are retained in the appendix; cross-round differences also
change partitions and teachers and cannot isolate the source-U intervention.

Five outers and ten paired draws use $N=100$ real target labels,
$M\in\{0,2000\}$ additional label-blind training queries, and fifteen or
fifty depth-two trees. Teachers are fixed 200-tree depth-six XGBoost,
200-tree depth-twelve RF with minimum leaf size five, and logistic
regression with $C=1$. The same three 100-record source tuning draws
and source development AUC select a shared tree base per capacity from
child weight $\{1,5,10\}$ and L2 $\{1,10,30\}$.
Each family selects $\alpha\in\{0,.25,.5,.75,1\}$ separately for each
query budget; choices are frozen before target evaluation. No target
validation labels or test-driven capacity choice are allowed.

\paragraph{Matched objectives and query policies.}
Eight families comprise native and staged raw/rank KD, native and staged
mapped-rank KD, and raw/rank listwise. For $K=N+M$, pointwise training uses
\[
(1-\alpha)\sum_{i\in L}\mathrm{BCE}(y_i,p_i)
+\alpha\frac{N}{K}\sum_{i\in L\cup U}\mathrm{BCE}(t_i,p_i),
\]
where $t_i$ is raw probability, empirical rank, or
$\sigma(4(r_i-.5))$. The last mapping matches rank-listwise teacher logits.
Listwise training substitutes $\alpha N\KL(Q\|\operatorname{softmax}(f))$
for the auxiliary term. Staged pointwise and listwise share five hard-label
warmup rounds, initialization, remaining rounds, and a numerical Hessian
floor of $10^{-6}$. Pointwise BCE and listwise KL still differ in loss
semantics and curvature; this is not a single-factor coupling intervention.
Native KD retains its weighted-soft-label intercept and full training schedule.

Four policies report source-selected $M=0$, source-selected $M=2000$,
$M=2000$ with the original $M=0$ alpha, and an $M=2000$ control permuting
only new unlabeled teacher scores. The fixed-alpha arm separates adding
queries from alpha reselection; source reselection need not improve target
performance. Shuffle retains labeled scores, feature inputs, and score
multisets. Alpha zero explicitly removes U, reproducing its corresponding
label-only student exactly instead of changing histogram cuts.

\paragraph{Probability-offset diagnostic and references.}
For every student, fit a scalar $b$ using the same $N$ training labels:
\[
b^*=\arg\min_b\sum_{i\in L}\mathrm{BCE}(y_i,\sigma(f_i+b))
+\frac{\lambda}{2}b^2.
\]
A common $\lambda\in\{0,1,10,100\}$ per capacity is selected by source
development logloss using three label-staged models, then frozen across
families and policies. Larger supervised models use the fifteen-tree
reference's lambda. This intercept-only diagnostic uses in-sample margins,
differs from temperature scaling~\cite{guo2017}, and does not guarantee
calibration. It consumes no additional labels and cannot change ranking.
Its deployment adds only a stored scalar, with no teacher dependency.

Native/staged label-only, larger CatBoost and XGBoost selected on the same
source draws, and source-small weight delivery are reported separately.
Source-small uses all source-fit labels; adding its offset also uses the
$N$ target labels. Weight delivery and scores-only access are different
information interfaces even at equal student capacity.
The batch completes 52,635 model-training calls and 30,360 separate scalar
offset optimizations across three tasks. Neither is a replication count.

\paragraph{Uncertainty and audits.}
Outer partitions overlap within fixed archives. We retain paired draw
intervals conditional on each outer and report every outer effect; counts
of positive configuration means do not establish population significance.
Independent audits recompute source-only preprocessing, verify row budgets
and disjointness, reproduce source alpha/lambda choices, and test exact
alpha-zero and equal-alpha policy equivalence. Analytic objective derivatives
and intercept stationarity are checked. Exported-draw metrics are recomputed
from saved predictions; full matrices include negative results.

\paragraph{Frozen new-archive extension.}
A protocol frozen before downloading the new data or inspecting model results
selects the UCI Polish Companies Bankruptcy three-year forecast
horizon~\cite{polish}. Its 64 financial ratios extend the study to enterprise
finance. Label-blind exact-profile deduplication retains 10,416 of 10,503
statements; missing firm identifiers prevent a guarantee of unique companies
across partitions. Five randomizations allocate source development 1,000,
tuning pool 300, source queries 2,000, target pool 2,200, test 1,500, and
the remaining 3,416 statements to source fitting. This is random label
scarcity, not chronological forecasting. Six families and both capacities
retain selected/fixed/query-shuffle policies and the same source-only grids.
The source-fit budget differs from the credit archives. All 50 target draws
are retained, including two with zero positives; the other draws have
one to ten positives and tests have 51--85 of 1,500. Hard-label initialization
uses Jeffreys-smoothed prevalence; single-class larger supervised fits use
declared constant controls. There is no label-based resampling or new
probability-offset fit.

The extension completes 13,315 training calls, including twelve explicit
single-class constant outputs, 7,600 independent metric-array checks,
and 760 guarded student reloads. During interrupted official transfers,
training used the completely received selected ZIP member after length and
CRC verification, as permitted by a pretraining transport supplement.
Before result interpretation, both current and legacy official ZIP downloads
finish with the same full-archive SHA256; every member passes CRC checking.
The selected member is byte-identical to the reconstruction, and independent
parsing reproduces every registered feature, label, row ID, and profile hash.
A separate source receipt preserves the original input and model registration.
The three hypotheses test the matched rank comparison, fixed-query gains,
and practical references, without changing the horizon or selection policy.

\paragraph{Pretrained reference and label-excluding absorption.}
After the new-archive primary results, a registered follow-up evaluates
TabICLv2~\cite{qu2026} on the same Polish partitions and draws.
This is seen-archive evidence, not a second untouched-data confirmation.
TabICL 2.0.2 loads the v2 checkpoint from the official
\texttt{jingang/TabICL} repository at a fixed revision; mirror transport
matches the published LFS SHA256. Code and model card specify BSD-3-Clause.
Authors report synthetic pretraining, without our independently establishing
contamination or model-development exclusion. Eight ensemble members use
local weights and no automatic downloads. The constructor requests four
CPU threads; version 2.0.2 internally overrides this setting during prediction
with the maximum of the machine's logical-core count and the request.
The historical effective prediction thread count was not recorded.
Direct inference conditions on all 100 target labels. Five label-blind folds
produce student targets: each 20-row fold is scored using the other 80
labels, so its labels are excluded from its teacher context. Single-class
contexts return declared Jeffreys-smoothed constants. The documented v2
pretraining context range starts at 300; both 80 and 100 are below it.

No extra target labels or unlabeled queries are used ($M=0$). Source
development's first 500 rows refit the shared base and alpha choices on
three source tuning draws; the remaining 500 select one of native labels,
staged labels, raw/rank KD, and raw/rank listwise. Original full-development
choices are not reused. All five outer choices freeze before target
evaluation. The main batch completes 1,500 compact fits and 390 teacher
context calls, not pretrained-weight optimization steps. A supplementary
control registered during computation before target-result interpretation
permutes scores only within each OOF fold, preserving the score multiset
and both donor/recipient label exclusions; it adds 400 compact fits and
no teacher queries. These comparisons change teacher information and query
budgets relative to the source-trained $M=2000$ experiment and cannot
isolate the benefit of pretraining.

\paragraph{Query coverage versus teacher context.}
A further protocol registered after the negative pretrained follow-up,
before new scores, separates two interventions on the seen Polish archive.
It preserves all partitions, source processors, teacher weights, ensemble
size, objectives, and depth-two fifteen/fifty-tree budgets. A single Nebula
V100 uses Torch 2.6.0/CUDA 12.8, without AMP, FA3, or TF32; the historical
CPU experiment used Torch 2.5.1. We regenerate the $N=100,M=0$ anchor rather
than treating hardware/runtime differences as exact teacher replication.
All source choices freeze before new target evaluation; no new router is fit.

Query coverage holds $N=100$ and five $C=80$ teacher contexts fixed.
Each labeled row receives its label-excluding fold score; each unlabeled
row receives the mean of the same five teachers. Original label-blind query
orders give nested $M=500,2000$. Selected alpha uses the first 500 source
development rows; fixed alpha reuses the regenerated $M=0$ choice.
Both policies retain a matched U-only shuffle. Labeled targets use one OOF
teacher while U targets use an ensemble mean, so coverage and this
aggregation difference are not isolated from one another. Moreover,
$M=2000$ requires 10,000 teacher row predictions for U, not 2,000.

The context intervention fixes every student's 500 labels and changes
teacher contexts to nested $C=80,240,400$ subsets of the same heldout-fold
complements. Selected alpha is refit for each context; an additional
sensitivity applies the C400 alpha to C80 and C240. C400 is within the
documented pretraining context range; C80/C240 are below it. A separate
nested $N=100,300,500$ label-budget series uses ordinary fivefold contexts
and refits source bases per N; its cross-N changes are not context-only
effects. Direct teacher references condition on all N labels and retain
weights and context at inference. Source tuning uses three draws: distinct
labels total 200--216, 300, and 862--876 at N100/N300/N500 per outer,
respectively, in addition to 500 development labels for base/alpha choices;
the reserved other 500 do not select a method. The larger source tuning
cost therefore remains part of the label-budget series.

\section{Completed Results}
\begin{table}[t]
\centering\small
\caption{Heldout source-query pool, XGBoost teacher, $N=100$, fifty depth-two trees; five outers and ten draws. Selected $M=2000$ retunes alpha on source; fixed reuses $M=0$ alpha. Staged and mapped controls match the label warmup and rank-logit representation. LC shows 2016.}
\label{tab:heldout}
\begin{tabular}{llrrr}
\toprule
Task & Family & $M=0$ & Selected & Fixed\\
\midrule
LC & Raw KD & 0.6697 & 0.6691 & 0.6691\\
LC & Staged raw & 0.6681 & 0.6649 & 0.6649\\
LC & Mapped stage & 0.6691 & 0.6696 & 0.6690\\
LC & Raw list & 0.6823 & 0.6841 & 0.6840\\
LC & Rank list & 0.6814 & 0.6825 & 0.6833\\
Amex & Raw KD & 0.8879 & 0.8965 & 0.8960\\
Amex & Staged raw & 0.8887 & 0.8969 & 0.8975\\
Amex & Mapped stage & 0.8893 & 0.8931 & 0.8928\\
Amex & Raw list & 0.8933 & 0.9076 & 0.9061\\
Amex & Rank list & 0.8954 & 0.9063 & 0.9062\\
UCI & Raw KD & 0.7591 & 0.7661 & 0.7675\\
UCI & Staged raw & 0.7583 & 0.7661 & 0.7664\\
UCI & Mapped stage & 0.7250 & 0.7435 & 0.7376\\
UCI & Raw list & 0.7600 & 0.7668 & 0.7657\\
UCI & Rank list & 0.7540 & 0.7666 & 0.7666\\
\bottomrule
\end{tabular}
\end{table}

\paragraph{The historical RPA failure guides the controls.}
Earlier shared-parameter and multi-teacher results reject generic RPA
superiority. In the latter matrix, rank KD exceeds RPA in all twenty-four
LC capacity--teacher--budget--cohort means. Removing teacher clipping alone
does not rescue it. Detached min--max normalization, sample-dependent
gradient scale, diagonal/clipped curvature, and schedule/intercept choices
interact; no tested single intervention consistently repairs the update.
These diagnoses motivate fixed-temperature candidates and matched controls,
rather than retaining RPA as the preferred algorithm. Their full results,
including MIL evidence, remain in the appendix.

Figure~\ref{fig:rpa-diagnosis} brings the recorded update traces and
matched comparisons together: a positive correspondence contrast does
not imply superiority over ordinary distillation. The trace ratios
describe the executed updates; they do not isolate a sufficient repair.

\begin{figure*}[!t]
\centering
\includegraphics[width=\textwidth]{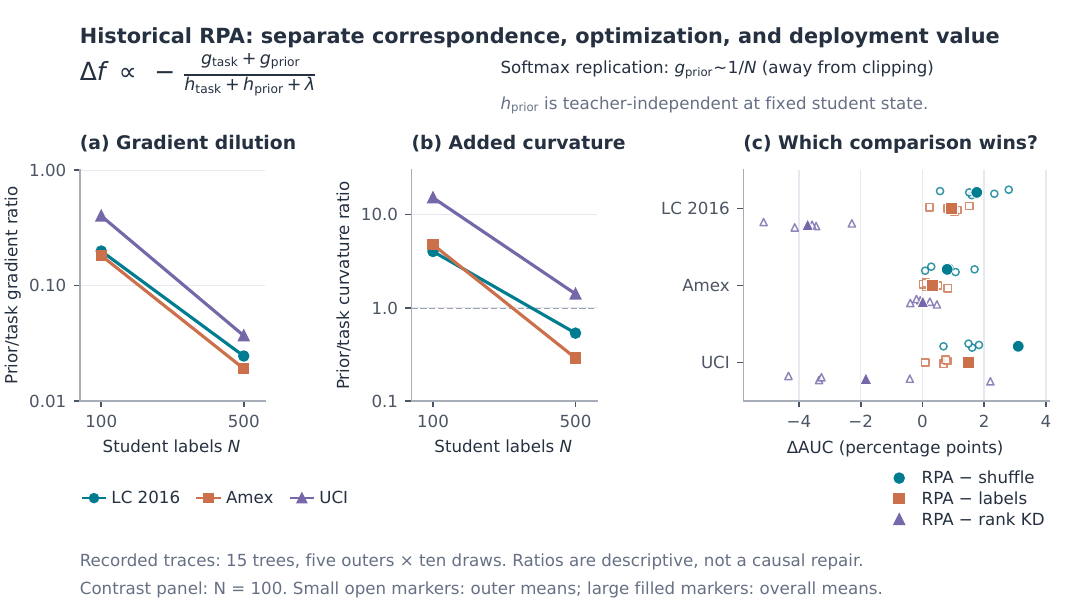}
\caption{Historical RPA diagnosis with source-trained XGBoost teachers
and cached numeric-full features: depth-two, fifteen-tree students on
all three credit archives. LC uses the first registered test cohort, 2016.
(a,b) Recorded prior/task gradient and curvature ratios average each
run's alignment-round summary over five overlapping outers and ten draws
at each $N$. The schematic Newton update highlights softmax dilution and
teacher-independent curvature at a fixed state; evolving-state traces
do not isolate a causal repair. (c) At $N=100$, RPA can exceed shuffled
scores and supervision while losing to native rank KD. Small open markers
retain all five outer effects; filled markers show their means. All three
contrasts use the same registered RPA arm. No population confidence
interval is implied.}
\label{fig:rpa-diagnosis}
\end{figure*}

\paragraph{Heldout queries help frequently, with explicit counterexamples.}
Table~\ref{tab:heldout} reports the latest protocol. Additional $M=2000$
queries with source alpha reselection improve the two listwise families
in forty-seven of forty-eight configuration means; holding $M=0$ alpha
fixed improves forty-six. These configurations span teachers, capacities,
tasks, and both LC cohorts and are correlated. In contrast to the preceding
query batch's forty-eight positive means, UCI now supplies counterexamples:
RF raw-listwise at fifteen trees declines by .00084 with reselection
and .00118 with fixed alpha; fixed-alpha logistic rank-listwise declines
by .00051. A query-coverage benefit is therefore bounded, not universal.

On Amex, all eight families improve
in each of their six teacher--capacity means under either query policy.
For fifty trees and XGBoost, raw-listwise changes from .8933 to .9076
with selection and .9061 with fixed alpha. On LC, both listwise families
improve in all twelve means each under both policies, while native raw KD
improves in only two of twelve. Native rank KD improves in two with
reselection but nine with fixed alpha. Source-selected alpha can therefore
reverse the apparent benefit of additional coverage on target evaluation.

\paragraph{The matched comparison survives, but does not select one universal loss.}
At selected $M=2000$, rank-listwise exceeds mapped staged pointwise KD
in all twenty-four teacher--capacity--task--cohort means, with positive
differences in every outer. Differences range from .0123 to .0284 on
LC, .0066 to .0132 on Amex, and .0228 to .0286 on UCI.
This persists with fixed alpha and at $M=0$ in aggregate means, so shared
warmup and numerical teacher mapping do not eliminate the result.
Conditional intervals can still include zero, and different objectives,
Hessian diagonals, and independently source-selected alpha remain involved.

Raw-listwise exceeds staged raw KD in twenty-one of twenty-four selected
means: twelve of twelve LC, five of six Amex, and four of six UCI.
Moreover, native raw KD exceeds rank-listwise in five of six UCI means.
UCI fifteen-tree XGBoost raw KD gives .7633 and rank-listwise .7529.
The meaningful boundary is the teacher--target representation and training
policy, not a claim that listwise training uniformly dominates pointwise KD.

\paragraph{Correspondence and interface value are distinct.}
Informative new scores exceed their U-only shuffle in all 192
family--teacher--capacity--task--cohort means. This supports correspondence
under equal feature and query budgets, but shuffle harm is not deployable
gain. The source-small weight-delivery reference exceeds every examined
selected-$M=2000$ scores-only mean in all eight task--capacity--cohort cells.
For fifty-tree LC 2016, Amex, and UCI, its AUCs are .6930, .9119, and
.7729, compared with the best examined scores-only means .6841, .9076,
and .7668. These best means summarize the matrix and are not a deployable
test-selected policy. On Amex, both displayed XGBoost-teacher listwise
capacities exceed the CatBoost reference (.8845); larger supervised
references are demanding controls, not evidence of exhaustive baseline coverage.

\begin{table}[t]
\centering\scriptsize
\caption{Raw and same-label offset-adjusted probability errors; XGBoost teacher, fifty depth-two trees, selected $M=2000$. LC evaluates 2016. Lower is better; AUC is unchanged.}
\label{tab:heldout-probability}
\begin{tabular}{llrrrr}
\toprule
Task & Family & Brier & Adjusted & Logloss & Adjusted\\
\midrule
LC & Rank KD & 0.2101 & 0.1881 & 0.6104 & 0.5606\\
LC & Raw list & 0.1731 & 0.1742 & 0.5251 & 0.5283\\
LC & Rank list & 0.1732 & 0.1745 & 0.5251 & 0.5287\\
Amex & Rank KD & 0.1646 & 0.1338 & 0.5062 & 0.4193\\
Amex & Raw list & 0.1213 & 0.1148 & 0.3769 & 0.3671\\
Amex & Rank list & 0.1310 & 0.1306 & 0.4102 & 0.4106\\
UCI & Rank KD & 0.1981 & 0.1702 & 0.5855 & 0.5205\\
UCI & Raw list & 0.1431 & 0.1423 & 0.4541 & 0.4529\\
UCI & Rank list & 0.1485 & 0.1490 & 0.4657 & 0.4671\\
\bottomrule
\end{tabular}
\end{table}

\paragraph{A scalar probability anchor is not a general repair.}
Table~\ref{tab:heldout-probability} reports probability errors.
The common-offset diagnostic preserves AUC. For selected native rank KD,
Brier and logloss improve in all twenty-four means: LC fifty-tree
XGBoost Brier changes from .2101 to .1881, but rank-listwise is .1732
before this adjustment. Rank KD's improvement is not ranking improvement
or proof of adequate final calibration.
The same offset worsens both Brier and logloss for both LC listwise
families in all twelve means each. At fifty trees, raw-listwise Brier
changes from .1731 to .1742 and logloss from .5251 to .5283.
On Amex, raw-listwise Brier improves in three of six means, whereas
rank-listwise logloss worsens in all six. UCI rank-listwise worsens
in both metrics in all six means. Common source-selected lambda and
in-sample target margins limit this diagnostic; a global offset does not
resolve slope, local probability error, or source-to-target selection mismatch.
We retain both outputs and do not choose between them using target tests.

\begin{table*}[t]
\centering\small
\caption{Frozen new-archive test: Polish bankruptcy, $N=100$, five overlapping outers and ten draws. AUC means; raw KD, mapped staged KD, raw/rank listwise use source-selected $M=2000$ alpha. Fixed rank-listwise reuses $M=0$ alpha; $\Delta_U$ is fixed minus $M=0$. Full policies and probability metrics are retained in the reproduction results.}
\label{tab:fresh}
\begin{tabular}{lrrrrrrr}
\toprule
Teacher / trees & Raw KD & Mapped stage & Raw list & Rank list & Fixed rank & $\Delta_U$ & Rank AP\\
\midrule
XGB / 15 & 0.6546 & 0.6512 & 0.7044 & 0.7277 & 0.7258 & 0.0256 & 0.1192\\
RF / 15 & 0.6994 & 0.6511 & 0.7127 & 0.7079 & 0.7065 & 0.0060 & 0.1067\\
LR / 15 & 0.7078 & 0.6504 & 0.6547 & 0.7131 & 0.7115 & 0.0122 & 0.1065\\
XGB / 50 & 0.6649 & 0.6741 & 0.7263 & 0.7487 & 0.7445 & 0.0316 & 0.1353\\
RF / 50 & 0.7104 & 0.6742 & 0.7289 & 0.7259 & 0.7199 & 0.0120 & 0.1321\\
LR / 50 & 0.7150 & 0.6672 & 0.6470 & 0.7165 & 0.7189 & 0.0133 & 0.1184\\
\bottomrule
\end{tabular}
\end{table*}

\begin{table*}[t]
\centering\small
\caption{New-archive unadjusted probability metrics under selected $M=2000$; no new probability-offset fit. AP is higher-is-better; Brier and logloss are lower-is-better.}
\label{tab:freshprob}
\begin{tabular}{llrrrrrr}
\toprule
Trees & Teacher & KD AP & Rank AP & KD Brier & Rank Brier & KD logloss & Rank logloss\\
\midrule
15 & XGB & 0.0847 & 0.1192 & 0.0441 & 0.0441 & 0.1907 & 0.1904\\
15 & RF & 0.0955 & 0.1067 & 0.0439 & 0.0441 & 0.1855 & 0.1898\\
15 & LR & 0.0984 & 0.1065 & 0.0439 & 0.0441 & 0.1858 & 0.1903\\
50 & XGB & 0.0906 & 0.1353 & 0.0438 & 0.0434 & 0.1880 & 0.1834\\
50 & RF & 0.1020 & 0.1321 & 0.0436 & 0.0434 & 0.1828 & 0.1827\\
50 & LR & 0.1025 & 0.1184 & 0.0437 & 0.0436 & 0.1835 & 0.1841\\
\bottomrule
\end{tabular}
\end{table*}

\paragraph{New-archive confirmation and its negative boundaries.}
Table~\ref{tab:fresh} reports every teacher--capacity configuration of the
previously unexamined bankruptcy task. Rank-listwise exceeds mapped staged
KD in six of six means with selected alpha and six with fixed alpha, with
positive differences in every outer. Selected differences range from .0493
to .0765, although four of thirty conditional outer draw intervals include
zero; two do so under fixed alpha. This confirms the bounded matched
comparison on one new archive, not an independent-institution effect.
Rank-listwise also exceeds native raw KD in all six selected-query means,
but the fifty-tree logistic difference is only .00155 and reverses in three
outers. Native rank KD is marginally stronger for fifteen-tree RF (.7082
versus .7079). These results do not identify a universally preferred loss.

Additional queries improve both listwise families in ten of twelve means
with selection and ten with fixed alpha. The two failures are logistic
raw-listwise: selected changes are $-.03147$ and $-.04291$ at fifteen and
fifty trees, negative in every outer. Fixed changes are $-.00669$ and
$-.01122$; reselection worsens these means by an additional .02477 and
.03169. Both rank-listwise capacities improve for every teacher under
either policy. Informative U scores exceed shuffled scores in all twelve
listwise means, including the two configurations with negative query value.
Thus correspondence and practical coverage gains remain distinct.

Probability and retrieval measures qualify the AUC result
(Table~\ref{tab:freshprob}). For fifty-tree XGBoost, rank-listwise AUC .7487
exceeds raw-listwise .7263, yet AP .1353 is below .2254. Against native raw
KD, rank-listwise improves AP in all six means but lowers Brier in only
three and logloss in three; small raw error differences do not establish
calibration. Native rank KD has Brier .1676--.2131, despite competitive
ranking. The low bankruptcy prevalence makes probability assessment essential.

Source-small weight delivery gives AUC .7495/.7827 at fifteen/fifty trees,
above every examined selected-query scores-only mean at the same nominal
capacity. Its AP is .1811/.2467; fifteen-tree XGBoost raw-listwise AP .1840
slightly exceeds that reference, so the AUC ordering is not universal across
metrics. This reference uses 3,416 source-fit labels and a different interface.
The original source-selected CatBoost gives .6498 AUC; the larger XGBoost
grid degenerates to constant predictions (.5000), already visible in its
source-development scores. These are retained references, not evidence that
strong supervised tabular methods have been exhausted.

An explicitly post-primary sensitivity broadens minimum child Hessian weight
to $\{.1,1,5\}$ and L2 to $\{1,10\}$ at depths three/six and 200 trees.
It selects from twelve candidates using the same source draws/development
records, freezes all choices before new target evaluation, and uses the same
100 target labels without teacher queries. The extra 230 calls are separate
from confirmation. Forty-seven of fifty target predictions are nonconstant;
AUC/AP/Brier/logloss are .6544/.0909/.0444/.1926. Each of the six primary
rank-listwise means exceeds it, with positive differences in every outer.
Five additional guarded reloads reproduce its exported students exactly.
This sensitivity addresses a degenerate reference and is not new confirmation
or a modern pretrained-baseline comparison.

\begin{table*}[t]
\centering\small
\caption{Seen-archive TabICLv2 follow-up on Polish bankruptcy, $N=100$, $M=0$: five overlapping outers and ten paired draws. OOF scores exclude each row's label. Source-only base and alpha choices are refitted; method selection uses a separate 500-row development half. Direct TabICLv2 retains pretrained weights and 100 context labels. AP is higher-is-better; Brier and logloss are lower-is-better.}
\label{tab:tabicl}
\begin{tabular}{llrrrr}
\toprule
Trees & Family & AUC & AP & Brier & Logloss\\
\midrule
15 & Native labels & 0.6381 & 0.0829 & 0.0443 & 0.1925\\
15 & Staged labels & 0.6416 & 0.0839 & 0.0443 & 0.1925\\
15 & Raw KD & 0.6340 & 0.0785 & 0.0447 & 0.1939\\
15 & Rank KD & 0.6386 & 0.0840 & 0.1166 & 0.3861\\
15 & Raw list & 0.6355 & 0.0811 & 0.0442 & 0.1917\\
15 & Rank list & 0.6394 & 0.0830 & 0.0443 & 0.1921\\
15 & Source selector & 0.6328 & 0.0791 & 0.0882 & 0.3019\\
50 & Native labels & 0.6459 & 0.0866 & 0.0442 & 0.1891\\
50 & Staged labels & 0.6437 & 0.0876 & 0.0441 & 0.1889\\
50 & Raw KD & 0.6384 & 0.0830 & 0.0444 & 0.1903\\
50 & Rank KD & 0.6355 & 0.0853 & 0.1784 & 0.5214\\
50 & Raw list & 0.6383 & 0.0814 & 0.0442 & 0.1902\\
50 & Rank list & 0.6364 & 0.0826 & 0.0440 & 0.1893\\
50 & Source selector & 0.6395 & 0.0807 & 0.0443 & 0.1903\\
Direct & TabICLv2 & 0.6542 & 0.1076 & 0.0487 & 0.2002\\
\bottomrule
\end{tabular}
\end{table*}

\begin{table}[t]
\centering\scriptsize
\caption{Source-only routing: AUC gains over native labels. Choices freeze before target evaluation; outer partitions overlap.}
\label{tab:tabicl-routing}
\begin{tabular}{rrlrr}
\toprule
Trees & Outer & Chosen family & Source gain & Target gain\\
\midrule
15 & 0 & Raw KD & 0.0525 & -0.0241\\
15 & 1 & Native labels & 0.0000 & 0.0000\\
15 & 2 & Raw KD & 0.0172 & -0.0308\\
15 & 3 & Rank KD & 0.0559 & -0.0114\\
15 & 4 & Rank list & 0.0443 & 0.0401\\
50 & 0 & Raw KD & 0.0429 & -0.0292\\
50 & 1 & Staged labels & 0.0027 & -0.0096\\
50 & 2 & Staged labels & 0.0055 & -0.0008\\
50 & 3 & Staged labels & 0.0231 & 0.0022\\
50 & 4 & Raw list & 0.0307 & 0.0052\\
\bottomrule
\end{tabular}
\end{table}

\paragraph{A modern teacher does not ensure useful absorption.}
Table~\ref{tab:tabicl} reports every family in the registered follow-up.
Direct TabICLv2 has AUC .6542, AP .1076, Brier .0487, and logloss .2002.
Relative to native labels for KD and staged labels for listwise, only
one of eight matched mean AUC differences is positive: fifteen-tree
rank KD gains .00053. The other differences range from $-.01041$ to
$-.00216$. Native rank KD also worsens probability errors: its Brier
is .1166/.1784 at fifteen/fifty trees, compared with .0443/.0442 for
native labels. This tests one specific low-context configuration and
does not establish a general failure of pretrained tabular models.

All eight informative-minus-fold-shuffle AUC means are positive
(.0379--.0706), but this does not rescue the matched deployment comparison.
Individual outer effects include ties and one negative contrast. Twenty-three
of 390 teacher contexts are single-class constants (seven source, sixteen
target). Rare labels, subrange teacher context, OOF 80-label versus direct
100-label context, and query coverage restricted to the 100 labeled rows
are plausible boundaries, not experimentally isolated causes. Capacity and
probability error also remain distinct explanations to test prospectively.

\paragraph{Pooled OOF scores have a rare-label comparability limit.}
A posthoc training-label-only diagnostic examines fold-specific class
prevalence; it does not refit a model or select a deployment method.
For a feature-free teacher returning its context's Jeffreys-smoothed rate,
one positive among 100 records yields .5/81 in its heldout fold and 1.5/81
in the other four folds. With twenty rows per fold, its pooled OOF AUC is
$19/198\approx .096$, despite valid label exclusion. Actual fifteen-source-
episode mean pooled teacher AUC is .5951 and mean within-fold concordance
is .6829; fourteen episodes have defined AUC. For the target training draws,
the corresponding means are .5441 and .6199 over 48 defined episodes;
the other two have no positive labels. A context-prevalence-only null gives
mean pooled AUC .2677/.2577 on source/target training draws.
Within-fold concordance weights eligible folds by positive--negative pairs;
the two summaries compare different pair sets and are not a causal effect.
These observations qualify pooled OOF AUC as a teacher-quality diagnostic.
They neither show that the actual teacher is feature-free nor establish
that fold calibration would repair deployed performance.

\paragraph{Source validation does not reliably predict target policy value.}
Table~\ref{tab:tabicl-routing} reports every frozen choice. Nine of ten
outer--capacity selections predict positive source AUC gain; six reverse
sign on target. Three gains survive and one selection falls back to native
labels. The selector's mean AUC is .6328/.6395, below native labels by
.00523/.00644. Separate development halves prevent direct reuse of method
selection labels for base/alpha tuning, but do not guarantee transfer of
the selected policy. This negative test identifies a transfer limitation:
transferability must be predicted and evaluated under a new frozen protocol,
rather than inferred from source validation alone.

\paragraph{Deployment compression is verified separately from accuracy.}
Independent audits check all 650 main and 400 control metric arrays,
reconstruct all 390 context row/feature/label hashes against recorded call
counts, and verify label exclusion and all source choices. Those hashes are
independently reconstructed receipts, not hashes captured at original fits.
All 100 compact and five direct exports reload with zero prediction error
and zero guarded external-access attempts. Main compact weight files occupy
11,084--45,328 bytes; direct assets with pretrained weights and context
occupy 110,435,720--110,435,774 bytes, excluding source preprocessing in
both ranges. Median warm prediction times for 128 rows are 0.1029 ms for
the 60 main compact exports and 1,359.5962 ms for the five direct exports.
Weights are already loaded; timings use two repeats under the recorded
implementation, whose thread override prevents a controlled four-thread
comparison. They exclude preprocessing, cold starts, and service overhead, and do not
establish per-request latency or a business SLA. Direct local ICL also
satisfies zero external calls, while retaining weights and labeled context;
only the compact route removes the teacher and context from deployment.

\paragraph{Query coverage changes the matched mean, with substantial uncertainty.}
Table~\ref{tab:boundary-query} reports the complete query intervention.
At both M500 and M2000, all eight family--capacity mean AUC contrasts
are positive against M0, their matched U-only shuffle, and matched
supervision, under both selected and fixed alpha. Selected-policy gains
over supervision range from .00189 to .00909 at M500 and .00135 to .00812
at M2000; corresponding fixed-policy ranges are .00109--.01132 and
.00073--.00836. Thus the descriptive improvement does not require alpha
reselection. It also does not establish stable practical gain: only 22/40
and 23/40 selected-policy outer--configuration effects are positive at
M500/M2000 (19/40 and 20/40 fixed). In each of the four comparisons, zero of forty
conditional paired-draw intervals lie wholly above zero. Draw intervals
condition on each overlapping outer and do not supply population inference.
Going from M500 to M2000 increases selected mean AUC in only four of eight
configurations; neither monotone query benefits nor a universal loss winner
is supported. Native rank KD still has Brier .1145/.1359 and logloss
.3808/.4240 at selected M2000, compared with native supervision's
.0443/.0442 and .1925/.1891. Ranking improvements do not resolve probability
estimation.

\begin{figure*}[!t]
\centering
\includegraphics[width=\textwidth]{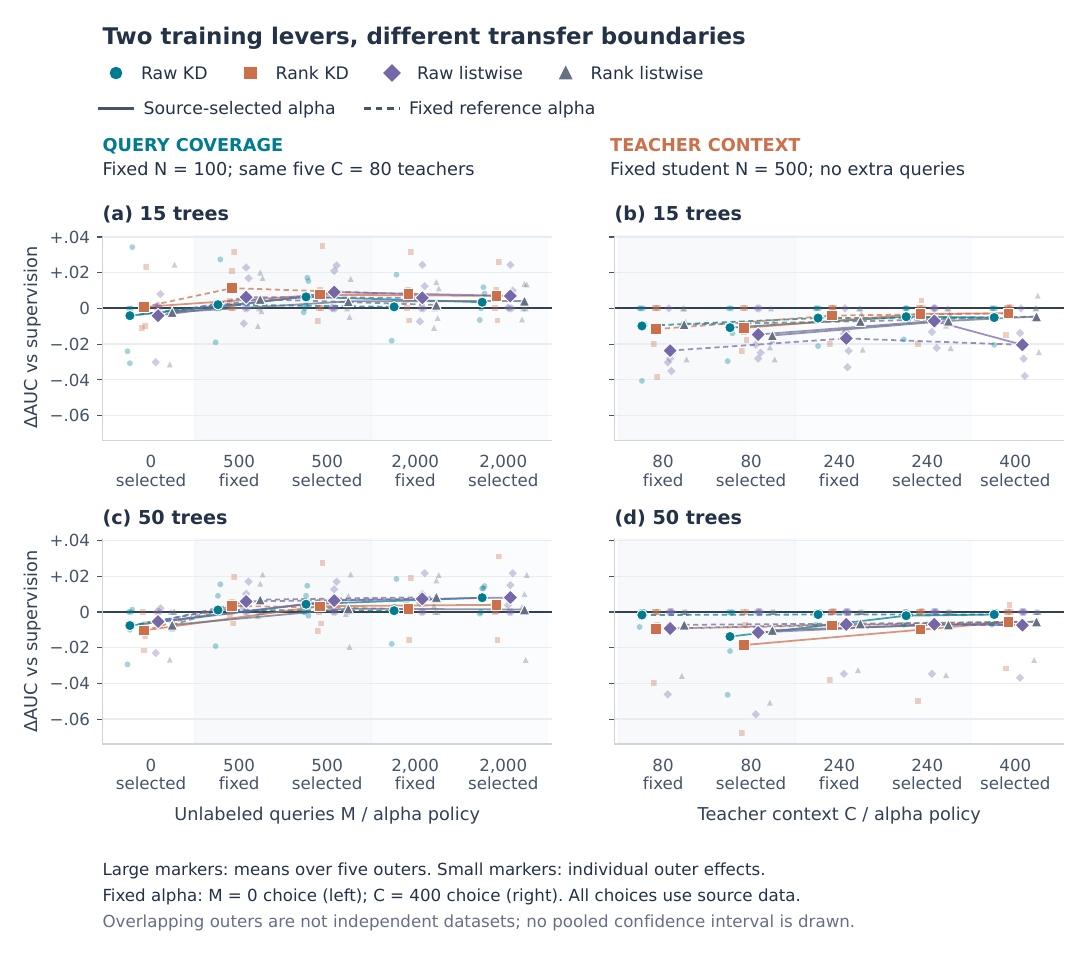}
\caption{Query coverage and teacher-context interventions on the seen
Polish archive. Large markers and curves show paired mean AUC changes
over five overlapping outers; small markers retain every outer effect.
KD is compared with native supervision, listwise with staged supervision.
Solid curves reselect alpha using source data; dashed curves hold the
$M=0$ choice (left) or $C=400$ choice (right). Query means turn positive
at $M=500/2000$ under both policies, but none of their corresponding
conditional supervised-contrast intervals lies wholly above zero; full
intervals remain in the saved artifacts. Every selected $N=500,C=400$
matched mean remains negative. Query scores average five teachers for U
but use one OOF teacher for labeled rows. The two interventions use
different student-label budgets and should not be compared as equal-cost
policies. These outers are not independent datasets; no pooled interval
is drawn.}
\label{fig:query-context-boundaries}
\end{figure*}

Figure~\ref{fig:query-context-boundaries} displays every outer effect,
so the positive aggregate query means can be read alongside their
negative and zero effects rather than mistaken for uniform benefits.

\begin{table*}[t]
\centering\scriptsize
\caption{Query coverage with fixed $N=100$ labels and five $C=80$ teachers. U scores average the same teachers; $M=500$ is nested in 2000. F/S denote fixed original/source-reselected alpha. AP/Brier/logloss show selected $M=2000$. Five overlapping outers and ten paired draws; all families retained.}
\label{tab:boundary-query}
\begin{tabular}{llrrrrrrrr}
\toprule
Trees & Family & $M=0$ & 500 F & 500 S & 2000 F & 2000 S & AP & Brier & Logloss\\
\midrule
15 & Raw KD & 0.6340 & 0.6401 & 0.6445 & 0.6389 & 0.6415 & 0.0810 & 0.0446 & 0.1950\\
15 & Rank KD & 0.6389 & 0.6494 & 0.6457 & 0.6461 & 0.6450 & 0.0820 & 0.1145 & 0.3808\\
15 & Raw list & 0.6375 & 0.6478 & 0.6506 & 0.6474 & 0.6485 & 0.0852 & 0.0442 & 0.1916\\
15 & Rank list & 0.6394 & 0.6467 & 0.6455 & 0.6434 & 0.6457 & 0.0847 & 0.0442 & 0.1916\\
50 & Raw KD & 0.6383 & 0.6470 & 0.6503 & 0.6467 & 0.6539 & 0.0886 & 0.0443 & 0.1910\\
50 & Rank KD & 0.6356 & 0.6494 & 0.6492 & 0.6475 & 0.6499 & 0.0872 & 0.1359 & 0.4240\\
50 & Raw list & 0.6383 & 0.6496 & 0.6501 & 0.6510 & 0.6518 & 0.0867 & 0.0442 & 0.1893\\
50 & Rank list & 0.6365 & 0.6505 & 0.6455 & 0.6520 & 0.6450 & 0.0843 & 0.0441 & 0.1890\\
\bottomrule
\end{tabular}
\end{table*}

\paragraph{Larger teacher context does not by itself rescue compact absorption.}
At fixed N500, increasing C80 to C400 improves seven of eight selected
mean AUCs and all eight fixed-C400-alpha means
(Table~\ref{tab:boundary-context}). Increasing C240 to C400 improves only
five selected and six fixed-alpha means; effects remain mixed across outers.
More importantly, every selected N500/C400 distillation mean loses to
matched supervision, by .00142--.02046 AUC. Fifteen-tree raw-listwise
falls from .7114 at C240 to .6981 at C400 under separately selected alpha;
with C400 alpha held fixed, the values are .7017 and .6981. This retains
a negative context boundary even without alpha reselection.

In the nested label series (Table~\ref{tab:boundary-label}), direct teacher
AUC rises from .6542 to .7674 and .8064 at N100/N300/N500. Student
supervision also strengthens, and no ordinary-context distillation family
exceeds its matched supervision at N300 or N500. This rules out treating
below-range context alone as a sufficient explanation for the earlier
negative configuration. It does not isolate whether limited student
capacity, target representation, fold comparability, or their interactions
cause the residual gap. Teacher improvements across N include extra real
labels and cannot be counted as gains at fixed label cost. Query coverage
at N500 has not been tested here; the context experiment uses M0.

\begin{table*}[t]
\centering\scriptsize
\caption{Teacher-context intervention at fixed student $N=500$, $M=0$. Label AUC is native for KD and staged for listwise. C80/C240/C400 independently source-select alpha; F400 applies the C400 alpha to each smaller context. AP/Brier/logloss show selected C400. Teacher contexts are nested subsets of identical heldout-fold complements.}
\label{tab:boundary-context}
\begin{tabular}{llrrrrrrrrr}
\toprule
Trees & Family & Label & C80 & C240 & C400 & C80 F400 & C240 F400 & AP & Brier & Logloss\\
\midrule
15 & Raw KD & 0.7173 & 0.7065 & 0.7126 & 0.7122 & 0.7075 & 0.7119 & 0.1171 & 0.0435 & 0.1821\\
15 & Rank KD & 0.7173 & 0.7064 & 0.7142 & 0.7145 & 0.7057 & 0.7133 & 0.1180 & 0.1081 & 0.3492\\
15 & Raw list & 0.7185 & 0.7038 & 0.7114 & 0.6981 & 0.6948 & 0.7017 & 0.1150 & 0.0433 & 0.1800\\
15 & Rank list & 0.7185 & 0.7033 & 0.7115 & 0.7139 & 0.7095 & 0.7113 & 0.1216 & 0.0434 & 0.1808\\
50 & Raw KD & 0.7403 & 0.7265 & 0.7382 & 0.7389 & 0.7386 & 0.7389 & 0.1470 & 0.0422 & 0.1714\\
50 & Rank KD & 0.7403 & 0.7218 & 0.7304 & 0.7348 & 0.7309 & 0.7328 & 0.1458 & 0.0562 & 0.2220\\
50 & Raw list & 0.7407 & 0.7293 & 0.7338 & 0.7334 & 0.7315 & 0.7338 & 0.1429 & 0.0424 & 0.1729\\
50 & Rank list & 0.7407 & 0.7306 & 0.7336 & 0.7353 & 0.7336 & 0.7342 & 0.1454 & 0.0424 & 0.1721\\
\bottomrule
\end{tabular}
\end{table*}

\begin{table*}[t]
\centering\scriptsize
\caption{Nested label budgets with ordinary fivefold teacher contexts: C80/C240/C400 for student N100/N300/N500. AUC; each N refits its shared source-only base. Direct ICL uses full N context and a larger deployment interface. Comparing across N changes both student and teacher labels and does not isolate context size.}
\label{tab:boundary-label}
\begin{tabular}{lrrrrrrrr}
\toprule
Trees & $N$ & Native & Staged & Raw KD & Rank KD & Raw list & Rank list & Direct ICL\\
\midrule
15 & 100 & 0.6381 & 0.6416 & 0.6340 & 0.6389 & 0.6375 & 0.6394 & 0.6542\\
15 & 300 & 0.6656 & 0.6660 & 0.6502 & 0.6617 & 0.6584 & 0.6614 & 0.7674\\
15 & 500 & 0.7173 & 0.7185 & 0.7122 & 0.7145 & 0.6981 & 0.7139 & 0.8064\\
50 & 100 & 0.6459 & 0.6437 & 0.6383 & 0.6356 & 0.6383 & 0.6365 & 0.6542\\
50 & 300 & 0.6764 & 0.6764 & 0.6587 & 0.6643 & 0.6654 & 0.6648 & 0.7674\\
50 & 500 & 0.7403 & 0.7407 & 0.7389 & 0.7348 & 0.7334 & 0.7353 & 0.8064\\
\bottomrule
\end{tabular}
\end{table*}

\paragraph{Boundary experiment integrity and deployment.}
The GPU stage records 1,820 context calls, including 26 declared
single-class constants, and captures actual context hashes at fits.
Independent checks verify 845 teacher arrays, context nesting, label
exclusion, all source choices, and all 6,750 metric arrays from 11,700
compact fits. An export filename collision overwrote earlier outer models;
we preserve the original registration and reconstruct 528 draw-zero
models using frozen scores, choices, and seeds. Every reconstructed file
matches its originally recorded SHA256 and all saved test predictions;
132 surviving files are copied byte-identically. Reconstruction is
additional execution work, not new experimental evidence. All 660 unique
compact exports then reload independently with zero prediction error and
zero guarded external attempts. We preserve the registered runner and
provide a separately versioned filename repair for future reproduction.
These checks establish a compact inference dependency path, not stable
accuracy gains or a prospective transfer predictor.

\begin{figure*}[!t]
\centering
\includegraphics[width=\textwidth]{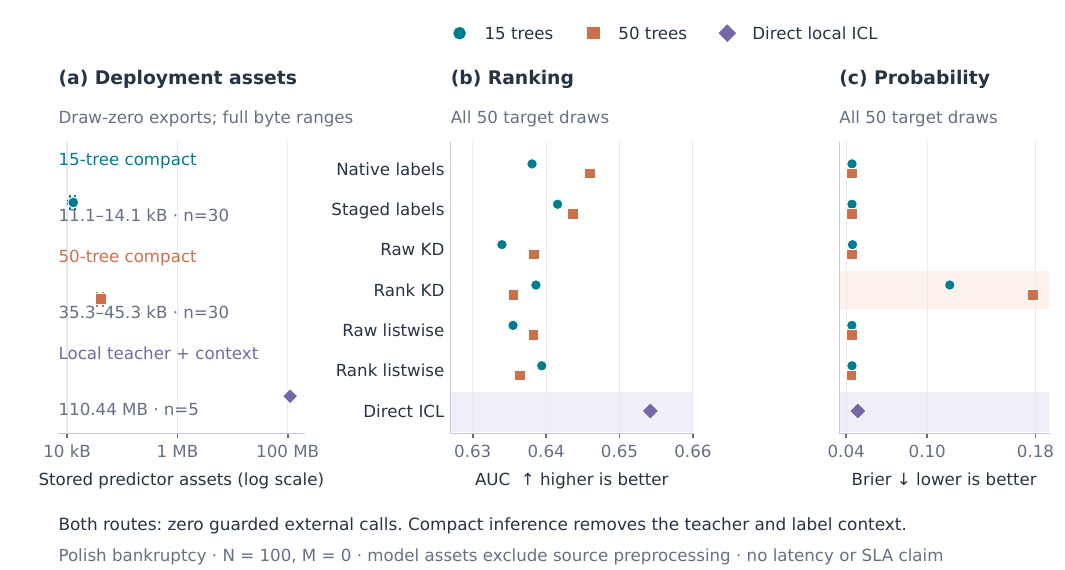}
\caption{Deployment footprint and predictive quality in the registered
TabICLv2 follow-up on the seen Polish archive ($N=100,M=0$).
(a) Asset ranges cover draw-zero exports from five outers: thirty compact
models per tree budget and five direct local teachers; source preprocessing
is excluded. (b,c) AUC and Brier are means over all fifty paired target
draws, not metrics restricted to those exports. Native and staged labels,
raw/rank KD, and raw/rank listwise are all retained. Direct ICL keeps
pretrained weights and all 100 context labels; compact absorption uses
label-excluding 80-label teachers during training and deploys student
weights with preprocessing. Both routes make zero guarded external calls.
The overlapping outers do not provide independent population inference.
No latency, service-SLA, or Pareto-optimality claim is made.}
\label{fig:deployment-quality}
\end{figure*}

Figure~\ref{fig:deployment-quality} separates compactness from accuracy
and probability error in the earlier $N=100,M=0$ pretrained follow-up.
Its byte ranges describe the exported subset; predictive means retain
every target draw. These are complementary summaries, not paired points
on an accuracy--memory frontier.

\paragraph{Independent inference.}
The 3,000 heldout-query students and 760 new-archive students, including
repeated/zero-alpha configurations, are reloaded with only weights, frozen source preprocessing, raw public
features, stored offsets, and reference predictions. No teacher or source
records are loaded. Maximum raw prediction error is below $6\times10^{-8}$
and adjusted error below $3\times10^{-8}$; bankruptcy reload errors are zero.
New-archive deployment assets including the source processor occupy
14,962--18,022 bytes at fifteen trees and 39,145--49,347 at fifty; actual
leaf counts range from 15--60 and 50--200 across all target draws and
source-small references. Nominal tree capacity does not imply equal leaves.
Python socket/DNS and subprocess
guards record zero attempted external accesses after imports; this is runtime
evidence, not operating-system isolation or a formal proof. Warm CPU timing
excludes service overhead and cold starts and establishes no business SLA.

\section{Limitations and Future Work}
The latest controls complete independent source-U selection, matched
larger-query schedules and mappings, fixed-alpha policies, and a common
probability-offset diagnostic on three examined credit archives. A protocol
frozen before downloading a previously unexamined enterprise-finance archive
then confirms the bounded matched comparison and retains query failures.
This is one new archive with overlapping random partitions, no firm IDs,
a smaller source-label budget, and explicit rare-label initialization;
it cannot establish chronological or cross-institution transfer.
We study one primary tree student family, conventional source teachers,
source-selected higher-capacity references, and one pretrained TabICLv2
model with both below-range and in-range context diagnostics. The query
intervention retains mixed outer effects; the seen-archive design and unequal
labeled/U target aggregation require a new prospective confirmation. A degenerate original XGBoost
reference is disclosed alongside a post-primary sparse-label sensitivity.
No result establishes an LLM-specific benefit, robustness across modern
pretrained tabular competitors, or production probability calibration.
The earlier MIL work covers a synthetic mechanism and one public natural
task; it cannot establish a universal supervision-granularity rule.

A central open question is whether source-observable quantities can predict
transfer boundaries under a new frozen evaluation. Future evaluations should compare
all-allowed features and a naturally unavailable-feature interface, additional modern
pretrained competitors and context-budget boundaries with documented provenance,
and stronger probability
estimators under the same real-label budget. Query and context effects have
now been separated within one model, but no source-observable boundary
rule or cross-archive modern-teacher confirmation has been established.
Additional natural MIL tasks are needed to test the generality of the
granularity hypothesis. The current experimental reproduction package
exists locally and covers the bankruptcy experiment,
its baseline sensitivity, pretrained follow-up, fold-local controls, and
query/context intervention with an export-recovery receipt. It has not yet
been publicly released, and earlier credit experiments are not yet included
in that standalone package. New-archive replication and fitted-model counts
do not substitute for a contribution beyond established distillation
and listwise learning or a complete public reproduction release.

LLM tabular priors such as TabLLM~\cite{hegselmann2023} motivate external
knowledge, but conventional source teachers do not verify their provenance
or advantage. Distillation theory~\cite{menon2021} and listwise learning
precede these objectives. Original industrial merchant-credit experiments
were already completed; configuration-specific findings remain historical
validation, not new public replications. Company data are not rerun here.

\section{Conclusion}
Training-only knowledge should be judged by deployable value at explicit
information and resource budgets. Correspondence benefits, an objective's
optimization behavior, and probability quality are separate requirements.
Heldout-query controls and a frozen new bankruptcy archive retain a bounded
rank-listwise advantage over mapped staged KD. Native raw KD still wins
most UCI credit comparisons, while additional queries harm logistic
raw-listwise on the new task under both policies. Source weight delivery
remains stronger in AUC under its different interface; AP and probability
errors qualify ranking advantages. The pretrained follow-up adds a negative boundary: useful correspondence
does not establish compact-model gain, and six of nine source-predicted
policy improvements reverse on target. Zero external calls are independently
verified deployment behavior, including for the locally hosted teacher.
A registered query intervention turns all eight matched mean differences
positive under selected and fixed alpha, but mixed outer effects and
conditional intervals prevent a stable-gain claim. Larger context alone
fails to beat supervision at N500 even as the direct teacher strengthens.
Open questions include predicting these boundaries
prospectively, validating query coverage on new data, improving probability
estimation, broadening modern references, and completing the reproduction release.

\clearpage
\appendix
\section{Earlier Registered Protocols}
These protocols are chronological controls on previously examined archives;
their counts and partition definitions differ from the main heldout-query batch.

\paragraph{Frozen tuning and capacity.}
The teacher is depth-six, 200-tree XGBoost. Principal students use fifteen
depth-two trees, learning rate .05, row subsampling .7, and column
subsampling .5. Fixed label-only uses L2=10 and minimum child weight=5.
Independent supervised tuning searches minimum child weight
$\{1,5,10\}$ by L2 $\{1,10,30\}$. RPA searches
$\gamma\in\{.1,.3,.8,1.5,3\}$ by $\tau\in\{.3,.5,1\}$;
each KD family searches $\alpha\in\{0,.25,.5,.75,1\}$.
Each family uses the same three source tuning draws and source development
AUC; choices remain fixed across target $N$. Unequal candidate counts are
disclosed. Actual leaves are recorded because equal depth/tree budgets need
not imply equal effective capacity.

Target acquisition samples rows uniformly before reading their labels,
with $N\in\{100,200,500\}$ and no additional target validation labels.
Five outer protocol randomizations each refit the teacher, source feature
selection, and tuning; twenty paired target draws share indices and model
seed across arms. Test samples are 6,000 customers/records per
evaluation partition. Pilot uses different seeds and is excluded.

\paragraph{Uncertainty and reproducibility.}
Outer randomizations reuse one fixed public archive and overlap in rows;
they are not independent institutions. We report individual outer effects
and paired draw bootstrap intervals conditional on each outer teacher and
test partition. A five-outer bootstrap is descriptive protocol variation,
not a population significance guarantee. Years, configurations, and fitted
model counts are not independent replications. Indices, overlap matrices,
source choices, predictions, optimizer traces, and export checks are saved.

\paragraph{Registered shared-parameter follow-up.}
After observing the first batch, we register a controlled follow-up using
the same source teachers, feature vocabularies, source tuning draws, target
draws, and test partitions. It is not a fresh-test confirmation. Within
each capacity, label-only source development selects child weight and L2
from the nine configurations above; every objective receives that same base.
Objective-specific weights are then source-selected and frozen across target
budgets. This controls base-parameter differences rather than providing
each method its independently optimal joint parameter search.
Capacities are depth two with fifteen or fifty trees and depth three with
fifty trees. Nine arms include label-only, staged label-only, raw/rank KD,
rank-KD shuffle, RPA, RPA shuffle, RPA curvature-only, and teacher-free
label smoothing with $t_i=(1-\epsilon)y_i+\epsilon/2$ and
$\epsilon\in\{0,.1,.25,.5,.75\}$. All capacities and budgets are reported;
we do not select a capacity using target-test performance. Actual leaves
remain distinct from the nominal tree budget. RPA retains teacher clipping
to $[.05,.85]$ and its warmup, while native KD uses unclipped probabilities;
the comparison still includes differences beyond target representation.

\paragraph{Multi-outer natural MIL follow-up.}
Five new customer-level randomizations split UCI into teacher fitting
(10,000), development (1,700), tuning pool (300), target pool (12,000),
and test (6,000). All groups are disjoint within an outer; outers overlap.
Signed-log amount transforms and standardization use source fitting only.
The shared student has a sixteen-unit encoder, attention pooling, and five
static context fields at the bag head. A logistic instance teacher inherits
customer labels: six rows per customer do not create six independent labels.
Its mean instance probability supplies bag KD and bag-ranking targets;
min--max-softmax instance probabilities supply attention-risk targets.
Three source tuning draws select common label-only epochs from
$\{20,40,80\}$ and then objective weights. Raw/rank KD use the earlier
alpha grid; bag-ranking, risk-attention, and uniform attention use
$\{0,.01,.05,.1,.3,.5,1\}$. Label smoothing uses the earlier epsilon grid.
Zero weight returns exactly to label-only training. All neural arms share
five hard-label warmup epochs and initialization. Fixed-.5 bag-ranking,
risk-attention, and uniform controls are retained beside tuned arms.
Each informative target has a shuffle at the same selected or fixed weight.
Bag shuffles act across customers; attention shuffles act within bags,
so their intervention strengths are not equivalent. Twenty paired draws
at each $N\in\{100,300,1000\}$ give 4,800 student fits. The experiment
measures natural bag prediction without relevance ground truth.

\paragraph{Registered RPA diagnosis and mapping control.}
We subsequently register nineteen arms on the same three public tree tasks,
using allowed features, five outers, ten previously sampled paired draws,
$N\in\{100,500\}$, and depth-two capacities of fifteen and fifty trees.
Shared base parameters and original KD/RPA weights reuse source selections.
Interventions remove only teacher clipping, replace its representation by
ranks, remove prior curvature, or multiply the rank-RPA weight by $N/100$;
they do not retune RPA intervention weights. Staged KD reuses native alpha;
a hard-label intercept control changes full rank KD initialization.
Each new listwise/pairwise family independently selects alpha from
$\{0,.25,.5,.75,1\}$ on the same three source draws and development set.
Matched shuffle targets retain alpha. The nineteen-arm matrix contains
11,400 student fits; recorded model counts are not independent replications.

After seeing these diagnosis results, we register an additional pointwise
mapping control using soft targets $\sigma(4(r_i-.5))$, matching the
listwise teacher logits. Native and staged variants independently select
source alpha, and a third staged variant reuses native alpha. This adds
1,800 fits. Both stages use previously examined archives/tests; the latter
is a result-motivated control, not an independent confirmation. The
ten-draw means should not be compared as new independent observations
against the earlier twenty-draw averages.

\paragraph{Multi-teacher and deployment-interface follow-up.}
A new frozen seed namespace samples source, development, tuning, target,
and test rows uniformly without inspecting labels. Budgets retain the
earlier source-fit/development/tuning counts. A source-fit processor
imputes nonfinite values by source medians, clips to source .5\%/99.5\%
quantiles, and standardizes by source means and deviations. All methods
share this processor and all allowed numeric fields. Thus cross-batch
changes cannot be attributed solely to teacher family.

Teachers are fixed depth-six 200-tree XGBoost, 200-tree random forests
with depth twelve and minimum leaf size five, and logistic regression
with $C=1$. Five outers, ten draws, $N\in\{100,500\}$, and depth-two
fifteen/fifty-tree students compare raw/rank KD, raw/rank listwise, RPA,
and matched rank-KD/rank-listwise/RPA shuffles. Each capacity shares
label-selected source base parameters; each teacher separately selects
objective weights using the same source-only grids. Teacher quality on
test records is an isolated diagnostic and never a selection criterion.

Larger target-supervised controls select CatBoost~\cite{prokhorenkova}
depth $\{4,6\}$ and L2 $\{3,10\}$ at 200 iterations, or XGBoost depth
$\{3,6\}$ and $\{100,200\}$ trees, using the same source tuning draws.
They use only $N$ target labels, without target early stopping. They are
higher-capacity references, not exhaustive optimized tabular benchmarks;
source selections at $N=100$ also remain fixed at $N=500$.
Separate weight-delivery references use all source-fit labels to train a
small student directly, or train all but five trees on source and append
five target-label trees. Total depth/tree budgets match the primary
students. The batch completes 27,525 training calls, including 15,600
main target fits, 600 adaptations, 600 stronger target fits, 60 source
weight models, 45 teachers, and 10,620 source-tuning fits. These counts
measure work performed, not independent statistical evidence.

\paragraph{Additional unlabeled offline queries.}
After inspecting the interface results, we freeze an additional protocol
that retains the same teachers, processors, paired target draws, and base
parameters at $N=100$. Uniformly sampled $M\in\{0,500,2000\}$ target
records exclude the labeled records; $M=500$ is nested in $M=2000$.
Their outcomes are never supplied to the learner. Four objective families
cross both capacities and all teachers. $M=0$ reuses frozen weights;
larger $M$ independently source-selects alpha on the original three
tuning draws and development set. Source-side unlabeled features come
from teacher-fit records, without their labels being passed to the student.
Those teacher scores are in-sample, which limits source-to-target tuning
validity; this is not cross-fitted selection.

For $K=N+M$, pointwise KD uses
\[
(1-\alpha)\sum_{i\in L}\mathrm{BCE}(y_i,p_i)
+\alpha\frac{N}{K}\sum_{i\in L\cup U}\mathrm{BCE}(q_i,p_i),
\]
with ranks replacing $q_i$ for the rank variant. Equivalent weighted
soft targets implement the loss. After label warmup, listwise uses
$(1-\alpha)\sum_{i\in L}\mathrm{BCE}+\alpha N\KL(Q\|P)$ on all
$K$ records, with masked task gradients. Zero alpha removes the
unlabeled rows explicitly so that they cannot alter histogram cuts.
The three-task batch completes 21,600 student/tuning fits and exactly
reproduces 4,800 earlier $M=0$ prediction arrays.
An additional control, registered during training before inspecting this
batch's target metrics, permutes only the $M=2000$ new teacher scores.
Labeled-record scores, all feature rows, score multisets, source alpha,
initialization, and capacity remain fixed. It adds 3,600 fits and tests
new correspondence while retaining the same histogram inputs. It is
still a follow-up on seen archives, not a fresh external confirmation.

\section{Earlier Diagnostic and Interface Results}
\begin{table*}[t]
\centering\small
\caption{New frozen protocol on seen public archives, $N=100$, fifteen depth-two trees, five outers and ten draws. Same source-fitted preprocessing and label-selected base parameters within each capacity. Teacher families are independently source-fitted; all objective weights are source-selected. LC shows 2016; full budgets, capacities, cohorts and matched shuffles remain in the complete matrix.}
\label{tab:internalization}
\begin{tabular}{llrrrrrr}
\toprule
Task & Teacher & Label & Raw KD & Rank KD & Raw list & Rank list & RPA\\
\midrule
LC & XGB & 0.6231 & 0.6654 & 0.6714 & 0.6696 & 0.6710 & 0.6414\\
LC & RF & 0.6231 & 0.6622 & 0.6692 & 0.6711 & 0.6702 & 0.6449\\
LC & LR & 0.6231 & 0.6611 & 0.6716 & 0.6695 & 0.6715 & 0.6432\\
Amex & XGB & 0.8700 & 0.8757 & 0.8737 & 0.8788 & 0.8802 & 0.8735\\
Amex & RF & 0.8700 & 0.8752 & 0.8734 & 0.8810 & 0.8795 & 0.8731\\
Amex & LR & 0.8700 & 0.8750 & 0.8749 & 0.8748 & 0.8799 & 0.8727\\
UCI & XGB & 0.6899 & 0.7495 & 0.7201 & 0.7423 & 0.7431 & 0.7131\\
UCI & RF & 0.6899 & 0.7503 & 0.7158 & 0.7462 & 0.7462 & 0.7094\\
UCI & LR & 0.6899 & 0.7329 & 0.7063 & 0.7382 & 0.7355 & 0.7034\\
\bottomrule
\end{tabular}
\end{table*}


\begin{table*}[t]
\centering\small
\caption{UCI six-month MIL follow-up: mean AUC over five customer partitions and twenty paired draws. Same student and source-selected epochs within each outer. Tuned weights use source data only; fixed weights equal .5. Natural attention has no relevance ground truth.}
\label{tab:milrefit}
\begin{tabular}{rccccccc}
\toprule
$N$ & Label & Smooth & Rank KD & Bag rank & Attention & Fixed bag & Fixed attention\\
\midrule
100 & 0.7160 & 0.7135 & 0.7328 & 0.7340 & 0.7148 & 0.7341 & 0.7114\\
300 & 0.7350 & 0.7330 & 0.7373 & 0.7394 & 0.7318 & 0.7406 & 0.7256\\
1000 & 0.7419 & 0.7405 & 0.7389 & 0.7406 & 0.7379 & 0.7421 & 0.7296\\
\bottomrule
\end{tabular}
\end{table*}

\begin{table*}[t]
\centering\small
\caption{Exploratory RPA diagnosis on seen public archives, $N=100$: five outers and ten paired draws, with source-selected shared base parameters. Staged KD reuses native alpha. Fixed listwise and pairwise select alpha on source only. Mapped KD matches the fixed-listwise teacher logits through pointwise sigmoid targets; this control was registered after seeing diagnosis results. LC shows 2016; all cohorts and $N=500$ are retained in the complete matrix.}
\label{tab:diagnosis}
\begin{tabular}{llccccccc}
\toprule
Task & Trees & Rank KD & Staged KD & RPA & Rank RPA & Listwise & Mapped KD & Pairwise\\
\midrule
LC & 2x15 & 0.6830 & 0.6760 & 0.6459 & 0.6478 & 0.6791 & 0.6825 & 0.6767\\
LC & 2x50 & 0.6874 & 0.6837 & 0.6682 & 0.6740 & 0.6876 & 0.6868 & 0.6874\\
Amex & 2x15 & 0.8754 & 0.8747 & 0.8756 & 0.8746 & 0.8843 & 0.8752 & 0.8755\\
Amex & 2x50 & 0.8921 & 0.8909 & 0.8875 & 0.8869 & 0.8970 & 0.8914 & 0.8911\\
UCI & 2x15 & 0.7291 & 0.7046 & 0.7108 & 0.7060 & 0.7446 & 0.7296 & 0.7249\\
UCI & 2x50 & 0.7392 & 0.7171 & 0.7269 & 0.7183 & 0.7534 & 0.7345 & 0.7402\\
\bottomrule
\end{tabular}
\end{table*}

\paragraph{A larger shuffle gap need not mean a better student.}
With LendingClub all-numeric features, $N=500$, and 2016 evaluation,
size-normalized $\gamma$ increases the net shuffle difference from .0012
to .0112, but changes actual AUC only from .6775 to .6788. Under the
weak vocabulary, actual AUC decreases from .6217 to .6189. Thus the
size intervention does not consistently improve deployment performance.
Changing only the curvature floor has very small fitted effects in this
batch: the largest paired single-cell AUC difference is approximately
$5.7\times10^{-5}$. The earlier replicated-score diagnostic's high
floor activity cannot be assumed to describe actual fitted margins.

\paragraph{Replicated natural MIL and budget boundary.}
At $N=100$, mean AUC is .7160 for label-only, .73396 for source-tuned
bag ranking, .7328 for rank KD, and .7148 for source-tuned risk attention.
Bag ranking exceeds label-only and tuned attention in each of the five
outer randomizations. Its .0012 mean advantage over rank KD is small
and does not establish broad listwise superiority. Teacher-free smoothing
and tuned uniform attention give .7135 and .7159.
At $N=300$, bag ranking gives .7394 versus label-only .7350 and attention
.7318; its label comparison reverses in one outer. At $N=1000$, tuned
bag ranking gives .7406 versus label-only .7419, with four negative
outer label differences. The bag-ranking--attention comparison also
reverses in three outers despite a positive aggregate mean.

Fixed-.5 attention has lower mean AUC than fixed-.5 bag ranking at all
three budgets, but source tuning substantially reduces attention's loss.
One outer selects zero attention weight, making it exactly label-only;
the remaining selected weights are $1,.01,.05,.05$. Source tuning
therefore does not prove that positive attention supervision is safe.
At $N=100$, bag ranking's own shuffle falls to .5102: a large matched
shuffle difference includes severe shuffled-target harm and should not
be presented as the deployable gain, which is .0180 over label-only.
These results support a budget-dependent target-semantic distinction
for this teacher and architecture, not a universal granularity theorem.

\paragraph{Why the historical RPA update loses.}
Removing teacher clipping changes mean AUC by at most about .0014 in this
diagnosis and does not rescue RPA. Schedule and intercept differences
matter more, but do not explain all deficits. On LC, fifteen trees,
$N=100$, and 2016 evaluation, native rank KD gives .6830, staged rank KD
.6760, and historical RPA .6459. The schedule comparison accounts for
only part of that gap. UCI reverses the matched-staged comparison:
native rank KD gives .7291, staged rank KD .7046, RPA .7108, and full
rank KD with a hard-label intercept .6927. Initialization changes
curvature and the fitting path as well as the initial prediction.

Training traces also expose a scale imbalance. In the LC fifteen-tree
condition above, the mean per-round prior/task absolute-gradient ratio is
.1987 at $N=100$, versus .0245 at $N=500$; prior/task curvature ratios
are 4.0165 and .5341. Large curvature is not equivalent to a large
teacher-directed update. These traces support an optimization hypothesis,
not a unique causal explanation. Removing prior curvature helps rank RPA
in the short LC $N=100$ run (.6478 to .6539), but hurts the fifty-tree
counterpart (.6740 to .6660). Rank representation helps LC but hurts
the examined UCI comparisons; size scaling also lacks a universal benefit.
No single one of these interventions consistently repairs RPA.

\paragraph{A promising alternative with explicit limits.}
Table~\ref{tab:diagnosis} reports the low-budget diagnosis. On Amex with
fifteen trees and $N=100$, fixed rank-listwise gives .8843 versus native
rank KD .8754. Its five outer mean differences are positive, although one
conditional draw interval crosses zero. At $N=500$ the means are .8970
and .8907. Both fifty-tree conditions also have positive aggregate
differences, with outer reversals. The matched-mapping control does not
explain this improvement: at $N=100$, mapped native KD gives .8752 and
independently source-tuned mapped staged KD .8737. Fixed listwise remains
higher in all four Amex aggregate mapping comparisons. Thus a numerical
mapping alone is insufficient to account for this exploratory signal;
cross-record normalization and gradient/curvature remain intertwined.

This candidate is not a universal winner. LC gains over native rank KD
are inconsistent, pairwise distillation is mixed, and at UCI fifteen-tree
$N=100$ raw KD .7487 exceeds rank-listwise .7446. Many source selections
choose listwise alpha one, so its alignment phase contains no hard-label
task term; hard-label warmup still precedes it. Better discrimination also
does not guarantee better probability estimates: Amex fifteen-tree
$N=100$ listwise Brier/logloss are .1590/.4887 versus label-only
.1549/.4782. These results warrant teacher-family and new-protocol
confirmation rather than immediate algorithm-superiority claims.

\begin{table}[t]
\centering\small
\caption{Additional offline queries, XGBoost teacher, $N=100$, fifteen depth-two trees. Five outers and ten draws; each $M$ uses source-selected alpha. LC shows 2016. Larger $M$ changes the training-query budget, not the real-label budget.}
\label{tab:unlabeled}
\begin{tabular}{llrrr}
\toprule
Task & Objective & $M=0$ & $500$ & $2000$\\
\midrule
LC & Raw KD & 0.6654 & 0.6627 & 0.6613\\
LC & Rank KD & 0.6714 & 0.6689 & 0.6691\\
LC & Raw list & 0.6696 & 0.6732 & 0.6736\\
LC & Rank list & 0.6710 & 0.6734 & 0.6742\\
Amex & Raw KD & 0.8757 & 0.8837 & 0.8837\\
Amex & Rank KD & 0.8737 & 0.8772 & 0.8784\\
Amex & Raw list & 0.8788 & 0.8882 & 0.8895\\
Amex & Rank list & 0.8802 & 0.8874 & 0.8860\\
UCI & Raw KD & 0.7495 & 0.7527 & 0.7527\\
UCI & Rank KD & 0.7201 & 0.7375 & 0.7420\\
UCI & Raw list & 0.7423 & 0.7472 & 0.7480\\
UCI & Rank list & 0.7431 & 0.7528 & 0.7550\\
\bottomrule
\end{tabular}
\end{table}

\paragraph{Teacher-family replication and stronger references.}
Table~\ref{tab:internalization} reports the new frozen protocol.
RPA remains below rank KD in all twenty-four LendingClub
teacher--capacity--budget--cohort means. Rank-listwise minus rank KD
has mixed LC signs (nine positive means), while raw-listwise exceeds raw
KD in all twenty-four means; this comparison includes different warmup
and intercept schedules and is not a pure estimate of loss coupling.

On Amex, rank-listwise exceeds rank KD in ten of twelve
teacher--capacity--budget means. At fifteen trees and $N=100$, its AUCs
are .8802, .8795, and .8799 for XGBoost, RF, and logistic teachers,
versus rank KD .8737, .8734, and .8749. The outer effects are positive
in five, five, and four of the five randomizations respectively.
The result therefore extends beyond the original teacher without
establishing a universal advantage. At $N=100$, CatBoost gives .8841,
above these fifteen-tree candidates; at fifty trees rank-listwise gives
.8963, .8934, and .8946. At $N=500$, CatBoost gives .9102 and remains
a demanding reference. Higher tree count alone is not synonymous with
better learning from scarce labels.

On UCI, rank-listwise exceeds rank KD in eleven of twelve means, but
raw KD remains stronger for XGBoost and RF teachers in the fifteen-tree,
$N=100$ setting: .7495/.7503 versus rank-listwise .7431/.7462.
The logistic-teacher comparison reverses, .7329 versus .7355.
Amex fifteen-tree rank-listwise Brier scores are .1594/.1584/.1603
across teachers: favorable AUC does not itself establish calibrated
probabilities. Full probability metrics and negative configurations are
retained with the paired contrasts.

\paragraph{Delivering weights can be the simpler route.}
At fifteen trees and $N=100$, source-small AUCs are .6850 for LC 2016,
.8972 for Amex, and .7633 for UCI; corresponding five-tree target
adaptations of shorter source models give .6783, .8950, and .7570.
Across all sixteen task--capacity--budget--cohort means, source-small
exceeds every examined informative score-only method. This is not a
same-information algorithm superiority claim: weight delivery gives
the deployed student a function trained on all source-fit records,
whereas score-only students receive only $N$ record-level teacher outputs.
Both routes can satisfy zero external inference calls. The practical
choice consequently depends on provider capabilities as well as AUC.
Adaptation is not automatically beneficial under a small target budget.
This reference reallocates five trees from source fitting to target fitting;
its difference from source-small is not a pure adaptation effect at an
unchanged pretrained initialization.

\paragraph{Offline query coverage changes the comparison.}
Table~\ref{tab:unlabeled} keeps 100 real target labels and reports the
additional-query follow-up. With 2,000 new unlabeled queries, raw-listwise
and rank-listwise each improve in all twenty-four
teacher--capacity--task--cohort means; together these are forty-eight
correlated configurations, not independent datasets. On Amex with fifty
trees and the XGBoost teacher, raw-listwise AUC changes from .8932 to
.9067 and rank-listwise from .8963 to .9074; all five outer effects are
positive. On UCI, fifty-tree rank-listwise changes from .7522 to .7650,
also with five positive outer effects. Its Brier score improves from
.1550 to .1521, but calibration remains a separate goal.

Additional queries are not automatically beneficial. On LC 2016 with
fifteen trees, XGBoost-teacher raw KD changes from .6654 to .6613 and
rank KD from .6714 to .6691, while raw/rank listwise change from
.6696/.6710 to .6736/.6742. Native raw and rank KD each improve in
only four of twelve LC capacity--teacher--cohort means. These protocol
effects include alpha reselection and objective-specific schedules; they
do not identify a unique causal effect of list coupling or query count.

Permuting only the new unlabeled teacher scores produces lower mean AUC
than their informative counterparts in all ninety-six
task--teacher--capacity--cohort--objective configurations at $M=2000$.
Original labeled-record teacher scores and histogram inputs are retained.
The control supports useful new correspondence, but can also induce severe
harm: Amex fifty-tree rank-listwise with the XGBoost teacher gives .7596
under this shuffle, compared with .9074 informative and .8963 at $M=0$.
The .1477 shuffle gap must therefore be distinguished from the .0110
actual query-policy gain. The source-small reference remains higher at
.9120 under its separate weight-delivery interface. Extra training
coverage narrows this gap without erasing interface or cost differences.

\paragraph{Independent inference.}
Exported tree students are reloaded in a separate process reading only
student weights and feature arrays. Predictions match the original outputs
within $10^{-7}$ tolerance. Warmed single-record inference timing excludes
feature engineering, service overhead, and cold starts; no business SLA
is asserted. Neural students are also exported and reload-tested; UCI MIL
exports are verified without teacher or source files in a separate process.
The multi-outer neural reload allows $5\times10^{-7}$ absolute error
because FP32 matrix reductions change with batch size and CPU thread count;
the actual maximum error is retained for each export.
The diagnosis additionally reloads 570 exported trees and the mapping
control reloads ninety, reading no teacher or source files. Original
control predictions reproduce 4,800 corresponding prior-batch arrays
exactly; row hashes and source/target/test isolation are audited.
The multi-teacher batch independently verifies 870 exported students with
only frozen source preprocessing, raw input features, and student weights.
Python socket connections, DNS, datagram sending, and subprocess creation
are disabled during prediction; there are zero attempted external accesses.
This is a runtime guard, not operating-system network isolation or formal
verification. All predictions agree within $3\times10^{-8}$ absolute error.
Score-only model plus processor sizes range from 13,522 to 48,813 bytes.
Median array-to-probability CPU timings for warm batches of 128 records
are approximately .30--.34 milliseconds across the three tasks, with
concurrent workloads and no service/SLA claim. Raw-input parsing, cold
starts, and service overhead remain outside that measurement.
The additional-query batch and its new-record shuffle independently
verify another 720 exported students under the same guard. No teacher
or source records are read by these inference processes.

\section{Earlier Protocol Controls and MIL Pilots}
These completed precursor analyses are retained for auditing and mechanism context. Their means belong to their own protocols and are not independent confirmations of the main follow-ups.

\begin{table*}[t]
\centering\small
\caption{First batch: mean AUC across five source-refit protocols and twenty paired target draws per protocol. Label leaf regularization is independently tuned; KD/RPA retain historical base settings. Table~\ref{tab:shared} addresses this confound. No extra target validation labels. Allowed features are retained numeric LC/UCI fields or the Amex S/B vocabulary; weak fields use source labels only. All students have fifteen depth-two trees. Years are separate LC cohorts; H denotes a random customer holdout. Cells are not independent datasets.}
\label{tab:public}
\begin{tabular}{llrrccccc}
\toprule
Task & Features & $N$ & Test & Tuned label & Raw KD & Rank KD & Fixed RPA & Tuned RPA\\
\midrule
LC & Allowed & 100 & 2016 & .6393 & .6690 & .6741 & .6535 & .6531\\
LC & Allowed & 100 & 2017 & .6366 & .6651 & .6701 & .6505 & .6502\\
LC & Allowed & 200 & 2016 & .6528 & .6796 & .6830 & .6655 & .6648\\
LC & Allowed & 200 & 2017 & .6493 & .6748 & .6773 & .6610 & .6603\\
LC & Allowed & 500 & 2016 & .6757 & .6833 & .6873 & .6775 & .6773\\
LC & Allowed & 500 & 2017 & .6693 & .6774 & .6806 & .6712 & .6712\\
\addlinespace[2pt]
LC & Source weak & 100 & 2016 & .5887 & .5982 & .6108 & .6017 & .6012\\
LC & Source weak & 100 & 2017 & .5876 & .6011 & .6122 & .6018 & .6010\\
LC & Source weak & 200 & 2016 & .6018 & .6245 & .6335 & .6068 & .6061\\
LC & Source weak & 200 & 2017 & .5991 & .6210 & .6292 & .6056 & .6049\\
LC & Source weak & 500 & 2016 & .6213 & .6331 & .6364 & .6217 & .6214\\
LC & Source weak & 500 & 2017 & .6169 & .6269 & .6306 & .6173 & .6174\\
\addlinespace[2pt]
Amex & Allowed & 100 & H & .8715 & .8559 & .8540 & .8694 & .8711\\
Amex & Allowed & 200 & H & .8813 & .8769 & .8789 & .8779 & .8785\\
Amex & Allowed & 500 & H & .8905 & .8883 & .8874 & .8885 & .8884\\
\addlinespace[2pt]
Amex & Source weak & 100 & H & .8371 & .7795 & .8137 & .8148 & .8178\\
Amex & Source weak & 200 & H & .8532 & .8418 & .8505 & .8476 & .8458\\
Amex & Source weak & 500 & H & .8659 & .8605 & .8663 & .8626 & .8611\\
\addlinespace[2pt]
UCI & Allowed & 100 & H & .6979 & .6201 & .6359 & .6852 & .6834\\
UCI & Allowed & 200 & H & .7307 & .7268 & .7272 & .7241 & .7229\\
UCI & Allowed & 500 & H & .7523 & .7568 & .7470 & .7514 & .7514\\
\addlinespace[2pt]
UCI & Source weak & 100 & H & .6649 & .6211 & .6214 & .6700 & .6685\\
UCI & Source weak & 200 & H & .6964 & .6534 & .6704 & .6803 & .6822\\
UCI & Source weak & 500 & H & .7168 & .7240 & .7182 & .7185 & .7178\\
\addlinespace[2pt]
\bottomrule
\end{tabular}
\end{table*}

\begin{table*}[t]
\centering\small
\caption{Controlled follow-up on previously evaluated public archives: source-selected child weight and L2 are shared across objectives within each capacity. Mean AUC for all allowed features at $N=100$, five outers and twenty paired draws. LC uses the 2016 cohort here; all cohorts, budgets and feature regimes are retained in the complete result matrix. No test-based capacity selection.}
\label{tab:shared}
\begin{tabular}{llccccc}
\toprule
Task & Capacity & Label & Smoothing & Raw KD & Rank KD & RPA\\
\midrule
LC & 2x15 & 0.6393 & 0.6391 & 0.6802 & 0.6830 & 0.6494\\
LC & 2x50 & 0.6456 & 0.6454 & 0.6843 & 0.6869 & 0.6692\\
LC & 3x50 & 0.6438 & 0.6438 & 0.6833 & 0.6854 & 0.6681\\
Amex & 2x15 & 0.8715 & 0.8721 & 0.8760 & 0.8752 & 0.8738\\
Amex & 2x50 & 0.8838 & 0.8838 & 0.8893 & 0.8912 & 0.8866\\
Amex & 3x50 & 0.8829 & 0.8829 & 0.8890 & 0.8922 & 0.8862\\
UCI & 2x15 & 0.6979 & 0.6983 & 0.7478 & 0.7299 & 0.7129\\
UCI & 2x50 & 0.7000 & 0.7020 & 0.7565 & 0.7396 & 0.7282\\
UCI & 3x50 & 0.6986 & 0.6988 & 0.7527 & 0.7227 & 0.7372\\
\bottomrule
\end{tabular}
\end{table*}

\paragraph{Content transfer versus objective advantage.}
On LendingClub, tuned RPA has lower mean AUC than rank KD in all twelve
feature-by-budget-by-cohort cells. With all allowed numeric features,
$N=100$, and 2016 evaluation, supervised tuning gives .6393, tuned RPA
.6531, and rank KD .6741. The RPA--rank-KD differences for the five
outer randomizations are all negative; the mean is $-.0210$.
Fixed RPA minus its own shuffle is $+.0336$. These answer different
questions: the latter supports useful correspondence, while the former
rejects a superiority claim in this setting.

The all-feature results address a weakness of the original weak-vocabulary
study. They preserve the rank-KD advantage while increasing absolute AUC.
On Amex at $N=100$ with the allowed S/B vocabulary, tuned RPA exceeds
rank KD by .0171 but is approximately tied with independently tuned
supervision (.8711 versus .8715). With the source-weak vocabulary, tuned
RPA is below tuned supervision by .0193. Thus a favorable KD comparison
alone is insufficient when a stronger teacher-free control exists. The
first-batch supervised arm tunes leaf regularization while RPA tunes its
alignment parameters with historical base-tree settings. We address this
confound with the registered shared-parameter follow-up below.
Amex and UCI must be interpreted using their own matched protocols rather
than merging them into a universal effect. Table~\ref{tab:public} is generated
from completed results and never includes pending runs.

\paragraph{Shared-parameter capacity intervention.}
On LendingClub, the follow-up's fifteen-tree label predictions reproduce
the independently tuned first-batch label predictions exactly. Sharing
source-selected leaf regularization does not rescue RPA's comparison with
rank KD: RPA has lower mean AUC in all thirty-six
feature-by-capacity-by-budget-by-cohort cells. These are correlated
configurations, not thirty-six independent datasets. At $N=100$, all
allowed numeric features, and 2016 evaluation, label-only, RPA, and rank KD
give .6393, .6494, and .6830 with fifteen trees; the RPA--rank-KD
outer effects are negative in all five randomizations. With fifty
depth-two trees, corresponding AUCs are .6456, .6692, and .6869;
depth-three fifty-tree results are .6438, .6681, and .6854.
RPA exceeds label-only in thirty-four of the thirty-six cells, but that
does not establish an advantage over the stronger training-only competitor.
Teacher-free smoothing also remains near label-only in the displayed LC
cells. Increasing capacity narrows some objective gaps but does not
justify a general inverse-capacity rule.

Amex also reverses the first batch's favorable RPA--KD comparison:
rank KD has higher mean AUC in all eighteen follow-up cells. At $N=100$
with the allowed S/B vocabulary and fifteen trees, label-only gives .8715,
raw KD .8760, rank KD .8752, and RPA .8738. On UCI, all allowed features
and the same nominal budget, raw KD rises from the first batch's .6201
to .7478, rank KD rises from .6359 to .7299, and RPA gives .7129.
The fifteen-tree raw-KD mean leaf count rises from 29.93 to 49.37;
the intervention changes shared regularization and source-selected
mixing weights, so it is not a pure causal estimate of leaf count alone.
RPA exceeds rank KD in six of eighteen UCI cells; raw KD remains
the stronger displayed UCI comparator. These findings reject a generic
ordinal-over-probability claim as well as generic RPA superiority.
Table~\ref{tab:shared} displays all three prespecified capacities at
$N=100$ with allowed features; the complete machine-readable matrix
retains both feature regimes, all budgets, and both LC test cohorts.

\paragraph{Synthetic MIL is exploratory.}
We generate twelve-instance bags with one known signal location. Labels
are exactly $y=\mathbf 1\{\exists j:x_{j,0}>2\}$. Contextual distractors
agree with labels with probability .9 in source and target training;
evaluation uses agreement .9 or .5. A source logistic instance scorer
inherits bag labels as a proxy training target, and the mean of its
instance probabilities is the bag prior. All student arms use the same
16-unit attention network, initialization, samples, Adam learning rate,
and twenty epochs selected by source label-only development.

We compare label-only, bag raw/rank KD, across-bag ranking, instance-risk
attention, within-bag shuffle, uniform attention, and an oracle signal
target. Oracle targets are privileged mechanism bounds, not fair ordinary
teacher advantages. There are only three draws per $N\in\{100,300\}$.
At $N=300$ and context agreement .5, label-only AUC is .5588 and
risk-attention AUC is .5218, even though removing the top-attended signal
changes the true generated label in essentially every positive bag for
the risk-attention arm. Localization and prediction are therefore distinct
metrics in this mechanism. Limited performance even for the oracle arm
motivates epoch/capacity checks; these pilot results do not establish a
universal granularity law.

\paragraph{Natural UCI MIL pilot.}
We additionally use each customer's six months as instances, with five
static fields supplied to the bag head as context. Fixed signed-log transforms
and source-fit standardization precede training. A source logistic instance
scorer uses inherited bag labels, and its mean probability is the bag prior;
inheritance does not create additional independent labels. All seven student
arms use the same architecture, initialization, and eighty source-selected
epochs. At $N=100$, mean AUC over three draws is .7161 for label-only,
.7270 for bag rank KD, .7305 for across-bag ranking, and .7012 for
instance-risk attention. The matched instance shuffle and uniform controls
give .7092 and .7126. At $N=300$, corresponding label-only, bag-ranking,
and instance-risk AUCs are .7284, .7344, and .7085. Thus the two prior
targets have different effects for the same teacher and student in this
pilot. One source split, three draws, and a fixed auxiliary weight do not
establish a universal granularity condition; natural attention is not
ground-truth importance. We subsequently complete source-selected weights
and multi-outer replication under the follow-up protocol.


\begingroup\footnotesize
\begin{thebibliography}{10}
\setlength{\itemsep}{0pt}
\bibitem{guo2017} C. Guo, G. Pleiss, Y. Sun, and K. Q. Weinberger. On Calibration of Modern Neural Networks. In \emph{ICML}, PMLR 70:1321--1330, 2017.
\bibitem{hinton2015} G. Hinton, O. Vinyals, and J. Dean. Distilling the Knowledge in a Neural Network. arXiv:1503.02531, 2015.
\bibitem{cao2007} Z. Cao, T. Qin, T.-Y. Liu, M.-F. Tsai, and H. Li. Learning to Rank: From Pairwise Approach to Listwise Approach. In \emph{ICML}, pp. 129--136, 2007.
\bibitem{chen2016} T. Chen and C. Guestrin. XGBoost: A Scalable Tree Boosting System. In \emph{KDD}, pp. 785--794, 2016.
\bibitem{prokhorenkova} L. Prokhorenkova, G. Gusev, A. Vorobev, A. V. Dorogush, and A. Gulin. CatBoost: Unbiased Boosting with Categorical Features. \href{https://arxiv.org/abs/1706.09516}{arXiv:1706.09516}.
\bibitem{ilse2018} M. Ilse, J. M. Tomczak, and M. Welling. Attention-based Deep Multiple Instance Learning. In \emph{ICML}, PMLR 80:2127--2136, 2018.
\bibitem{hegselmann2023} S. Hegselmann et al. TabLLM: Few-shot Classification of Tabular Data with Large Language Models. In \emph{AISTATS}, PMLR 206:5549--5581, 2023.
\bibitem{menon2021} A. K. Menon et al. A Statistical Perspective on Distillation. In \emph{ICML}, PMLR 139:7632--7642, 2021.
\bibitem{amex} American Express. American Express -- Default Prediction. \href{https://www.kaggle.com/competitions/amex-default-prediction}{Kaggle competition}, 2022.
\bibitem{lendingclub} LendingClub Corporation. Lending Club Loan Data. \href{https://www.kaggle.com/datasets/adarshsng/lending-club-loan-data-csv}{Public Kaggle archive}.
\bibitem{uci} I.-C. Yeh. Default of Credit Card Clients. UCI Repository, 2009. \href{https://doi.org/10.24432/C55S3H}{doi:10.24432/C55S3H}.
\bibitem{qu2026} J. Qu, D. Holzm\"uller, G. Varoquaux, and M. Le Morvan. TabICLv2: A better, faster, scalable, and open tabular foundation model. \href{https://arxiv.org/abs/2602.11139}{arXiv:2602.11139}, 2026.
\bibitem{polish} S. Tomczak. Polish Companies Bankruptcy. UCI Machine Learning Repository, 2016. \href{https://doi.org/10.24432/C5F600}{doi:10.24432/C5F600}.
\end{thebibliography}
\endgroup
\end{document}